\documentclass[10pt,journal,compsoc]{IEEEtran}

\usepackage{amsmath}
\usepackage{diagbox}
\usepackage{graphics}
\usepackage{epsfig}
\usepackage{threeparttable}
\usepackage{xspace}
\usepackage{siunitx}
\usepackage{graphicx}
\usepackage{amssymb}
\usepackage{threeparttable}
\usepackage{color}
\usepackage[normalem]{ulem}
\usepackage{multirow}
\usepackage{float}
\usepackage{amsfonts}
\usepackage{bm}
\usepackage{array}
\usepackage[table]{xcolor}
\usepackage{colortbl}
\usepackage{diagbox}
\usepackage{rotating}
\usepackage{booktabs}
\usepackage{overpic}
\usepackage{enumitem}
\usepackage{colortbl}
\usepackage{algorithm}
\usepackage{algorithmic}
\usepackage{cite}
\usepackage{dblfloatfix}

\usepackage{math_symbols}
\usepackage{pifont}
\usepackage[caption=false,position=bottom]{subfig}

\usepackage[pagebackref=false,breaklinks=true,colorlinks,bookmarks=false]{hyperref}

\definecolor{mygray}{gray}{.9}
\definecolor{lightgray}{gray}{.95}
\definecolor{ggray}{RGB}{127,127,127}
\definecolor{reda}{RGB}{192,0,0}
\definecolor{redb}{RGB}{217,148,143}
\definecolor{myyellow}{RGB}{190,144,0}
\definecolor{mygreen}{RGB}{93,173,85}
\definecolor{myblue}{RGB}{30,90,100}
\definecolor{demphcolor}{RGB}{100,100,100}

\definecolor{datagreen}{RGB}{93,173,85}
\definecolor{datared}{RGB}{240,16,89}
\definecolor{datablue}{RGB}{0,114,188}

\newcommand{\grouptablestyle}[2]{\setlength{\tabcolsep}{#1}\renewcommand{\arraystretch}{#2}\centering\footnotesize}
\newcommand{\sgrouptablestyle}[2]{\setlength{\tabcolsep}{#1}\renewcommand{\arraystretch}{#2}\centering}
\newcolumntype{x}[1]{>{\centering\arraybackslash}p{#1pt}}
\newcolumntype{I}{!{\vrule width 1pt}}
\newcolumntype{d}[1]{>{\raggedright\arraybackslash}p{#1pt}}
\newcolumntype{b}[1]{>{\raggedleft\arraybackslash}p{#1pt}}
\newcommand{\hlg}[1]{\textcolor{mygreen}{#1}}

\newcommand{\bbetter}[4]{
    \sgrouptablestyle{1pt}{1}
    \begin{tabular}{b{#1}d{#2}}
    {#3} &
    {\fontsize{6.5pt}{1em}\selectfont \hlg{\textbf{$\uparrow$#4}}}
    \end{tabular}
}

\newcommand{\void}[3]{
    \sgrouptablestyle{1pt}{1}
    \begin{tabular}{b{#1}d{#2}}
    {#3} &
    ~
    \end{tabular}
}
\newcommand{\subt}[4]{
    \sgrouptablestyle{0pt}{1}
    \begin{tabular}{b{#1}d{#2}}
    {#3} & {#4}
    \end{tabular}
}

\makeatletter
\DeclareRobustCommand\onedot{\futurelet\@let@token\@onedot}
\def\@onedot{\ifx\@let@token.\else.\null\fi\xspace}

\def\eg{\textit{e.g}\onedot} 
\def\ie{\textit{i.e}\onedot} 
\def\cf{\textit{c.f}\onedot} 
\def\etc{\textit{etc}\onedot} 
\def\wrt{w.r.t\onedot} 
\def\etal{\textit{et al}\onedot}
\def\cf{\textit{cf}\onedot}
\makeatother

\makeatletter
\newcommand{\thickhline}{%
    \noalign {\ifnum 0=`}\fi \hrule height 1pt
    \futurelet \reserved@a \@xhline
}
\newcommand{\sub}[1]{{$_{\text{#1}}$}}
\makeatother

\makeatletter

\newcommand{\tablestyle}[4]{ 
    \centering
    \resizebox{#1\textwidth}{!}{
    \setlength\tabcolsep{#2pt}
    \renewcommand\arraystretch{#3}
    #4
    }
}
\makeatother

\begin{document}

\title{Local-Global Context Aware Transformer for Language-Guided Video Segmentation}

\author{Chen~Liang, Wenguan~Wang,~\IEEEmembership{Senior~Member,~IEEE,} Tianfei~Zhou, \\
        Jiaxu~Miao, Yawei~Luo, and~Yi~Yang,~\IEEEmembership{Senior~Member,~IEEE}
\IEEEcompsocitemizethanks{
\IEEEcompsocthanksitem C. Liang, W. Wang, J. Miao, Y. Luo, and Y. Yang are with ReLER, CCAI, Zhejiang University. (Email: \{leonliang, yangyics\}@zju.edu.cn, \{wenguanwang.ai,yaweiluo329\}@gmail.com, jiaxu.miao@yahoo.com)
\IEEEcompsocthanksitem T. Zhou is with ETH Zurich. (Email: ztfei.debug@gmail.com)
\IEEEcompsocthanksitem Corresponding author: Yi Yang
}
}

\markboth{IEEE TRANSACTIONS ON PATTERN ANALYSIS AND MACHINE INTELLIGENCE}%
{Liang \MakeLowercase{\textit{et al.}}: \textsc{Locater} Local-Global Context Aware Transformer for Language-Guided Video Segmentation}

\IEEEtitleabstractindextext{%
\begin{abstract}
    We explore the task of language-guided video segmentation (LVS). Previous algorithms mostly adopt 3D CNNs to learn video representation, struggling to capture long-term  context and easily suffering from visual-linguistic misalignment. In light of this, we present \textsc{Locater} (\underline{lo}cal-global \underline{c}ontext \underline{a}ware \underline{T}ransform\underline{er}), which augments the Transformer architecture with a finite memory so as to query the~entire video with the language expression in an efficient manner. The memory is designed to involve two components -- one for persistently preserving global video content, and one for dynamically gathering local temporal context and segmentation history. Based on the memorized local-global context and the particular content of each frame, \textsc{Locater} holistically and flexibly comprehends the expression as an adaptive query vector for each frame. The vector is used to query the corresponding frame for mask generation. The memory also allows \textsc{Locater} to process videos with linear time complexity and constant size memory, while Transformer-style self-attention computation scales quadratically with sequence length. To thoroughly examine the visual grounding capability of LVS models, we contribute a new LVS dataset, A2D-S$^+$, which is built upon A2D-S dataset but poses increased challenges in disambiguating among similar objects. Experiments on three LVS datasets and our A2D-S$^+$ show that \textsc{Locater} outperforms previous state-of-the-arts.
    Further, we won the 1\textit{st} place in the Referring Video Object Segmentation Track of the 3\textit{rd} Large-scale Video Object Segmentation Challenge, where \textsc{Locater} served as the foundation for the winning solution. Our code and dataset are available at: \url{https://github.com/leonnnop/Locater}.
\end{abstract}

}

\maketitle
\IEEEdisplaynontitleabstractindextext
\IEEEpeerreviewmaketitle

\IEEEraisesectionheading{\section{Introduction}\label{sec:introduction}}
\IEEEPARstart{L}{anguage-guided} video segmentation (LVS)~\cite{wang2021survey}, also known as language-queried video actor segmentation~\cite{gavrilyuk2018actor}, aims to segment a specific object/actor in a video referred by a linguistic phrase. LVS is a challenging task as it delivers high demands for understanding whole video content and diverse language concepts, and, most essentially, comprehending alignments between linguistic and spatio-temporal visual clues at pixel level. Till now, popular LVS solutions are mainly upon FCN-style architectures with different visual-linguistic information fusion modules, such as dynamic convolution~\cite{gavrilyuk2018actor}, cross-modal attention~\cite{wang2019asymmetric,seo2020urvos}.

\begin{figure}[t]
    \begin{center}
        \includegraphics[width=.98\linewidth]{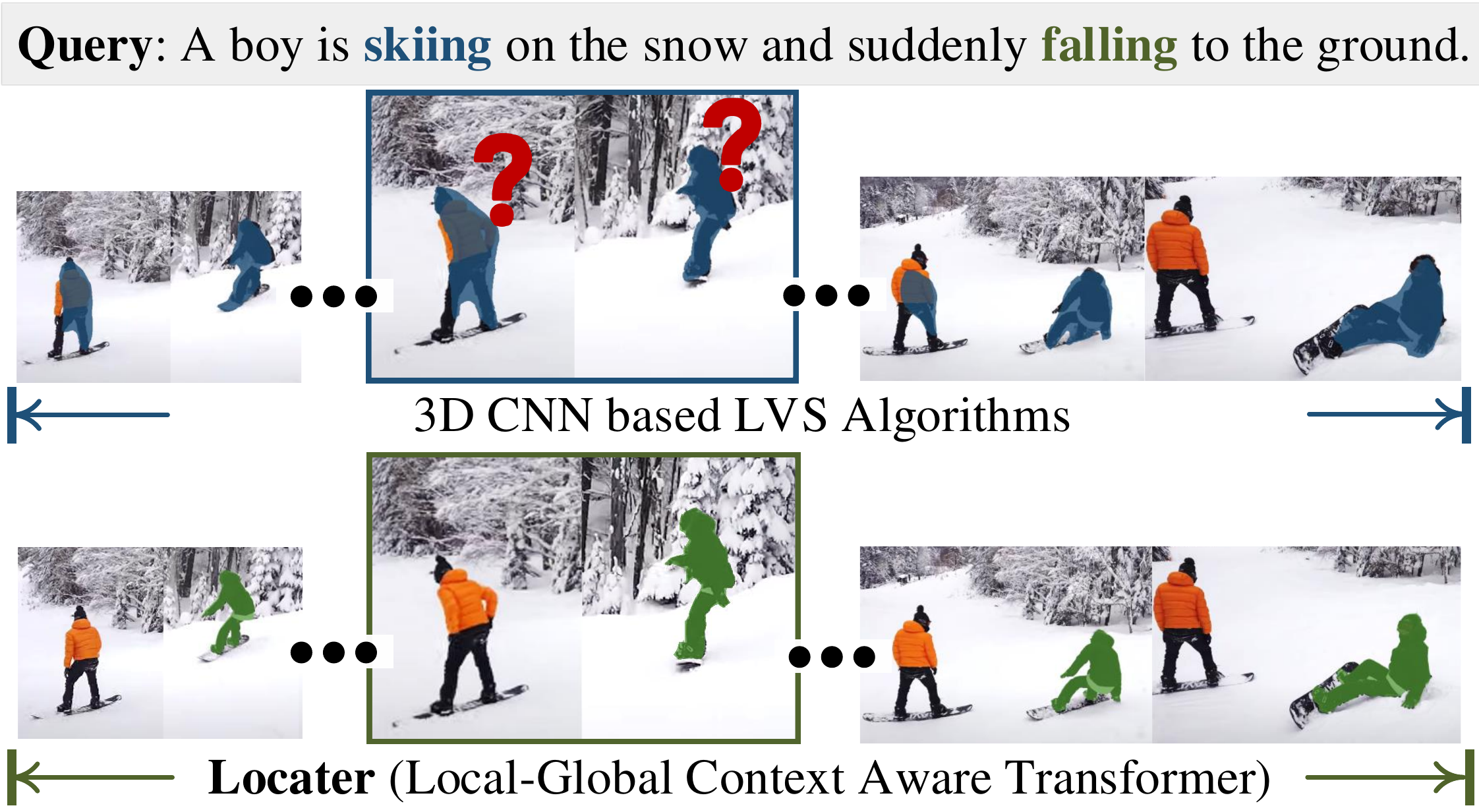}
    \end{center}
    \caption{\textbf{Our main idea.} {Previous LVS models are mainly built upon 3D CNNs, bounded by the local observations from short durations.
    They thus fail to identify the target referent in the early stage beyond the `fall' event.
    In contrast, \textsc{Locater} is fully aware of both local and global context, enabling a holistic understanding of entire video content and correctly grounding the phrase over the whole video.}}
    \label{fig:moti}
\end{figure}

Though impressive, most existing LVS solutions fail to fully exploit the global video context and the connection~between global video context and language descriptions. This is because they usually adopt 3D CNNs~\cite{carreira2017quo} for video representation learning. Due to the heavy computational cost and locality nature of 3D kernels, many algorithms can only gather limited information from short video clips (with 8-/16-frame length~\cite{ningpolar,gavrilyuk2018actor,wang2020context,wang2019asymmetric,mcintosh2020visual}) but discard long-term temporal context, which, however, is crucial for LVS task. First, due to the underlying \textit{low-level video processing} challenges such as object occlusion, appearance change and fast motion, it is hard to segment the target referent safely by only considering short-term temporal context~\cite{lu2019see,miao2021vspw,wang2020symbiotic}. Second, from a perspective of \textit{high-level visual-linguistic semantics understanding}, short-term temporal context is insufficient for reaching a complete understanding of video content and struggles for tackling complex activity related phrases.
As shown in Fig.~\ref{fig:moti}, given the description of ``a boy is skiing on the snow and suddenly falling $\cdots$'', the target skier cannot be identified unless with a holistic understanding of the video content, as the `fall' event happens in the second half of the video.

To cope with these temporal and cross-modal challenges posed in LVS task, we propose a new model, named as \textsc{Locater} (\underline{lo}cal-global \underline{c}ontext \underline{a}ware \underline{T}ransform\underline{er}). Building upon Transformer encoder-decoder architecture~\cite{vaswani2017attention}, \textsc{Locater} first utilizes self-attention to enhance per-frame representations with intra-frame visual context and linguistic features. To further incorporate temporal cues into per-frame representation, \textsc{Locater} builds an external, \textit{finite} memory, which encodes multi-temporal-scale context and also bases content retrieve on the attention operation. This makes \textsc{Locater} a fully attentional model, yet with greatly reduced space and computation complexities.

In particular, the external memory has two components: \textbf{i)} the former is to persistently memorize global temporal context, \ie, highly compact descriptors summarized from frames sampled over the entire span of a video; \textbf{ii)}  the latter is to online gather local temporal context and segmentation history, from past segmented frames. The global memory is maintained unchanged during the whole segmentation procedure while the local memory is dynamically updated with the segmentation processes. Hence, \textsc{Locater} gains a holistic understanding of video content and captures temporal coherence, leading to contextualized visual representation learning. Conditioned on the stored context and particular content of one frame, \textsc{Locater} vividly interprets the expression by adaptively attending to informative words, and forms an expressive query vector that specifically suits that frame. The specific query vector is then used to query corresponding contextualized visual feature for mask decoding.

With such a memory design, \textsc{Locater} is capable of comprehensively modeling temporal dependencies and cross-modal~interactions in LVS, and processing arbitrary length videos with low time complexity $\Ocal(N)$ and constant space cost $\Ocal(1)$. In contrast, Transformer-style self-attention requires to maintain all
$\Ocal(N^2)$ cross-frame dependencies. In addition, we introduce a deeply supervised learning strategy that feeds supervision signal into intermediate layers of \textsc{Locater}, for easing training and boosting performance.

We further notice that in A2D-S~\cite{gavrilyuk2018actor}, the most popular LVS dataset, a large portion of videos only contain very few but obvious objects/actors, making such task trivial.
To address this limitation, we synthesize A2D-S$^+$, a \textit{harder} dataset, from A2D-S.
Each video in A2D-S$^+$ is either selected or created to contain several semantically similar objects through a semi-automatic contrasting sampling~\cite{sadhu2020video} process.
It doubles the dataset difficulty in terms of the number of grounding-required examples per video, while with a negligible cost of human labour.

The contributions can be summarized into three folds:
\begin{itemize}[leftmargin=*]
	\setlength{\itemsep}{0pt}
	\setlength{\parsep}{1pt}
	\setlength{\parskip}{1pt}
	\setlength{\leftmargin}{-10pt}
\item We propose the pilot work that tackles LVS task with a memory augmented, fully attentional Transformer framework. Several essential designs (\ie, finite memory, progressive cross-modal fusion, contextualized query embedding, deeply supervision) significantly facilitate the network learning and finally lead to impressive performance.
\item The finite memory enables elegant long-term memorization and extraction of cross-modal context, while in the meantime, getting rid of the unaffordable space and computation cost brought by the quadratic complexity of the conventional attention in Transformers.
\item We alleviate the excess of trivial cases in the current most popular LVS benchmark, \ie, A2D-S, through introducing a \textit{harder} synthesized dataset with little human effort.
\end{itemize}

We empirically demonstrate that \textsc{Locater} consistently surpasses existing state-of-the-arts
across three popular ben- chmark datasets (\ie, \bpo{3.6}/\bpo{5.9}/\bpo{1.1} on A2D-S$_{\!}$~\cite{gavrilyuk2018actor} {$_{\!}$}(\S\ref{sec:a2d})/\\
\noindent J-HMDB-S~\cite{gavrilyuk2018actor} {$_{\!}$}(\S\ref{sec:jhmdb})/R-YTVOS~\cite{seo2020urvos} {$_{\!}$}(\S\ref{sec:urvos}) in mIoU/mIoU/ $\Jcal\&\Fcal$) and also performs robust on our challenging A2D-S$^+$ dataset (\S\ref{sec:contra_setting}).
Moreover, based on \textsc{Locater}, we ranked 1\textit{st} place in the Referring Video Object Segmentation (RVOS) track in the 3\textit{rd} Large-scale Video Object Segmentation Challenge$_{\!}$~\cite{vosc2021} (YTB-VOS\sub{21}), outperforming the runner-up by a large margin (\eg, \bpo{11.3} in $\Jcal\&\Fcal$; see \S\ref{sec:challenge}).
Further, we conduct a series of diagnostic experiments on several variants of \textsc{Locater}, verifying both the efficacy and efficiency of our core model designs (\S\ref{sec:ablation}).

\section{Related Work}
To offer necessary background, we review literature in~LVS (\S\ref{sec:rLVS}), and discuss relevant work in other areas (\S\ref{sec:rRIS}-\ref{sec:external_mem}).

\subsection{Language-guided Video Segmentation (LVS)}\label{sec:rLVS}
Studies of LVS are initiated by~\cite{gavrilyuk2018actor,khoreva2018video}. With the theme of efficiently capturing the multi-modal nature of the task, existing efforts investigate different visual-linguistic embedding schemes, such as capsule routing~\cite{mcintosh2020visual}, dynamic convolution~\cite{gavrilyuk2018actor,wang2020context,ningpolar}, and cross-modal attention~\cite{khoreva2018video,wang2019asymmetric,hui2021collaborative}. Owning to the difficulty in handling variable duration of videos, existing algorithms are mainly built on 3D CNNs, suffering from an intrinsic limitation in modeling dependencies among distant frames~\cite{varol2017long,piergiovanni2018learning,yang2021hierarchical,yang2021multiple,wang2022towards}. To remedy this issue, we exploit self-attention to gather cross-modal and temporal cues in a holistic and efficient manner. This is achieved by augmenting Transformer architecture with an explicit memory, which gathers and stores both global and local temporal context with learnable operations. Although~\cite{seo2020urvos} also enjoys the advantage of the outside memory, it does not consider global video context and requires multi-round complicated inference with a \textit{heuristic} memory update rule. In essence, we formulate the task in an encoder-decoder attention framework, instead of following the widely-used CNN-style architecture. Two concurrent works~\cite{wu2022language,botach2022end}, inspired by the query-based paradigm in object detection~\cite{carion2020end} and instance segmentation~\cite{wang2021end}, formulate LVS as a sequence prediction problem and introduce Transformers for querying object sequences. Though effective, they have to translate the whole video as input tokens. In contrast, with the aid of the external memory, \textsc{Locater} yields comparable performance with improved efficiency.

\subsection{Referring Image Segmentation (RIS)}\label{sec:rRIS}
As the counterpart of LVS in image domain, RIS has longer research history, dating back to the work~\cite{hu2016segmentation} of Hu and others in 2016. Primitive solutions directly fuse concatenated language and image features with a segmentation network to infer the referent mask~\cite{hu2016segmentation,liu2017recurrent,shi2018key,margffoy2018dynamic,li2018referring}. More recent approaches
focus on designing modules to promote visual-linguistic interactions, \eg, cross-modal attention~\cite{liang2021rethinking,ye2019cross}, progressive dual-modal encoding~\cite{feng2021encoder}, linguistic structure~\cite{hui2020linguistic,huang2020referring,yang2021bottom}.

\subsection{Referring Expression Comprehension (REC)}\label{sec:rREC}
REC is to localize linguistic phrases in images (phase grounding) or videos (language-guided object tracking, LOT), in a form of bounding box. The majority of studies, to date, follow a \textit{two-stage} procedure to select the best-matching region from a set of
bounding box proposals~\cite{hu2017modeling,hu2016natural,luo2017comprehension,yu2016modeling,yu2017joint,liu2019improving,bajaj2019g3raphground,wang2019neighbourhood}. Despite their promising results, the performance of two-stage methods is capped by the speed and accuracy of the proposal generator. Alternatively, a few recent works resort to a \textit{single-stage} paradigm, which embeds linguistic features into one-stage detectors, and directly predicts the bounding box~\cite{yang2019fast,liao2020real,luo2020multi}. Compared with REC, LVS is more challenging since it requires pixel-level joint video-language understanding. As for LOT, it is an emerging research domain~\cite{li2017tracking}. Existing solutions typically equip famous trackers with a language grounding module~\cite{li2017tracking,feng2021siamese,wang2021towards}. Note that there are many differences between LOT and LVS: i) they explain phrases with different visual constituents (pixel \textit{vs} tight  bounding box); ii) LOT adopts first-frame oriented phrases due to its tracking nature, while LVS is more aware of video understanding, \ie, considering arbitrary objects and unconstrained referring expressions~\cite{khoreva2018video}.

\subsection{Semi-automatic{ $_{\!}$}Video{ $_{\!}$}Object{ $_{\!}$}Segmentation{ $_{\!}$}(SVOS)}\label{sec:rSVOS}
SVOS aims to selectively segment video objects based on first-frame masks. LVS also has a close connection with SVOS, as it replaces the expensive pixel-level intervention with easily acquired linguistic guidance. Depending on the utilization of test-time supervision, current SVOS models can be categorized into three classes~\cite{wang2021survey}: i) \textit{online fine-tuning based} methods first train a generic segmentation network and then fine-tune it with the given masks~\cite{DBLP:conf/cvpr/CaellesMPLCG17,voigtlaender2017online}; ii) \textit{propagation based} methods use the previous frame mask to infer the current one~\cite{DBLP:conf/cvpr/PerazziKBSS17,DBLP:conf/cvpr/JampaniGG17,yangdecoupling,yang2021associating,wang2018semi}; and iii) \textit{matching based} methods classify each pixel's label according to its similarity to the annotated target~\cite{DBLP:conf/iccv/YoonRKLSK17,wang2020paying,lu2021segmenting}. Although some recent matching based SVOS approaches~\cite{lu2021segmenting,yang2020collaborative,lu2020video} also leverage memory to reuse past segmentation information, we focus on a multi-modal task, and enhance Transformer self-attention with a fixed size memory to address the efficiency issue in modeling intra- and inter-modal long-term dependencies.

\subsection{Transformer in Vision-Language Tasks}\label{sec:rTVLT}
The remarkable successes of Transformer~\cite{vaswani2017attention} in NLP spur increasing efforts
applying Transformer for vision-language tasks, \eg, video captioning~\cite{lei2020mart}, text-to-visual retrieval~\cite{gabeur2020multi,Miech_2021_CVPR}, visual question answering~\cite{khan2020mmft}, and temporal language localization~\cite{zhang2021multi}. For generating coherent captions, \cite{lei2020mart} also utilizes memory to better summarize history information. However, our target task, LVS, requires fine-grained grounding on the spatio-temporal visual space, and our \textsc{Locater} exploits local and global video context as well as segmentation history in a comprehensive and efficient manner. Transformer-based \textit{pretrained} models also showed great potential in joint image-language embedding~\cite{lu2019vilbert,tan2019lxmert,li2020unicoder,chen2019uniter,su2019vl, hu2020iterative,murahari2020large,hao2020towards}. Some other efforts were made towards unsupervised video-language representation learning~\cite{sun2019videobert,zhu2020actbert}, while they still often utilize 3D CNNs for video content encoding.

\subsection{Transformer in Visual Referring Tasks} \label{sec:rTVRT}
Concurrent to our work, only a handful of attempts apply Transformer-style models for LVS~\cite{wu2022language,botach2022end}, RIS~\cite{ding2021vision}, and REC~\cite{li2021referring,deng2021transvg,qin2023coarse,kamath2021mdetr}. Compared with these efforts, our approach is unique in several aspects, including memory design, language-guided progressive encoding, contextualized query embedding, and deep supervision strategy.

\subsection{Neural Networks with External Memory} \label{sec:external_mem}
Modeling structures within sequential data has long been considered a foundational problem in machine learning. Recurrent networks (RNNs)~\cite{rumelhart1986learning}, \eg, LSTM~\cite{hochreiter1997long}, GRU~\cite{cho2014properties}, as a traditional class of methods, have been shown Turing-Complete~\cite{siegelmann1995computational} and successfully applied into a wide range of relevant fields, \eg, machine translation~\cite{sutskever2014sequence}, and video recognition~\cite{nicolicioiu2019recurrent,li2022locality}. However, due to the limited capacity of the latent network states, they struggle for long-term context modeling. To address this limitation, researchers~\cite{graves2014neural} enrich RNNs' dynamic storage updates with external memory design, whose long-term memorization/reasoning with explicit storage manipulations earns remarkable success in robot control~\cite{oh2016control,parisotto2017neural,wang2021structured}, language modeling~\cite{bahdanau2014neural,weston2014memory,sukhbaatar2015end}, \etc.

Drawing inspiration from these novel designs, we equip \textsc{Locater} with an external memory. This allows for persistent and precise information storage, as well as flexible manipulations using learnable read/write operations. Hence such memory-augmented architecture well supports long-term modeling, which is critical for LVS.

\section{Methodology}\label{sec:method}
Given a video $\Ical\!\!=\!\!\{I_{t\!}\!\in\!\!\RR^{W_{\!} \x_{\!} H_{\!} \x_{\!} 3} \}_{t=1\!\!}^{N_t}$ with $N_t$ frames and a language expression $E\!\!=\!\!\{e_w\}_{w=1\!}^{N_w}$ with $N_w$ words, LVS is to predict a sequence of masks $\{S_{t\!}\!\in\!\!\{0,1\}^{W_{\!} \x_{\!} H} \}_{t=1\!}^{N_t}$ that determine the referent in each frame. Our \textsc{Locater} can \textit{efficiently} query the \textit{entire} video with the expression, by exploiting global and local temporal context through an \linebreak external memory. We first review the preliminary for Transformer (\S\ref{sec:preliminary}), and then elaborate our model design (\S\ref{sec:Locater}).

\begin{figure*}[t]
    \begin{center}
        \includegraphics[width=1\linewidth]{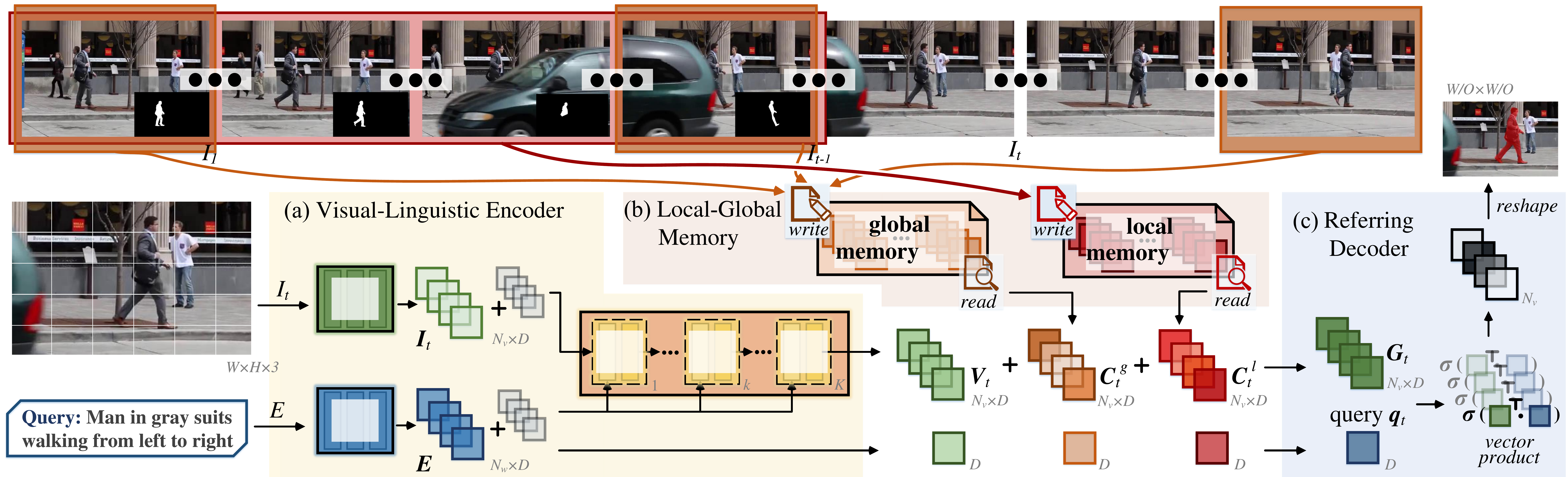}
        \put(-60, 105){\scriptsize$I_{N_t}$}
        \put(-110, 10){\scriptsize$\tilde{\Vbm}_{t}$}
        \put(-153.5, 10){\scriptsize$\tilde{\mbm}^{g}$}
        \put(-197, 10){\scriptsize$\tilde{\mbm}_{t}^{l}$}
        \put(-129, 72){\scriptsize$\Mcal^l$}
        \put(-212.5, 72){\scriptsize$\Mcal^g$}
        \put(-350, 72){\scriptsize$[\bm{\varepsilon}_{v_{\!}}]^{\!N_v}$}
        \put(-350, 31){\scriptsize$[\bm{\varepsilon}_{w_{\!}}]^{\!N_w}$}
        \put(-405, 56){\small$\Ecal_v$}
        \put(-406, 15.5){\small$\Ecal_w$}
        \put(-321, 39){\scriptsize$\Fcal^{1}_{vw}$}
        \put(-290, 39){\scriptsize$\Fcal^{k}_{vw}$}
        \put(-259.5, 39){\scriptsize$\Fcal^{K}_{vw}$}
        \put(-18, 73){\scriptsize$\hat{S}_{t}$}
        \put(-412, 77){\scriptsize (\S\ref{sec:3.2.1})}
        \put(-300, 68){\scriptsize (\S\ref{sec:3.2.2})}
        \put(-78, 64){\scriptsize (\S\ref{sec:3.2.3})}
    \end{center}
    \caption{\textbf{Illustration of our \textsc{Locater}.} {Building upon Transformer encoder-decoder architecture, \textsc{Locater} maintains a finite memory to collect and retain both global and local temporal context. By referencing the memory, \textsc{Locater} can reach a holistic understanding of video content and query the entire video with the expression, with linear computation complexity and constant storage consumption. See \S\ref{sec:Locater} for details.}}
    \label{fig:pipe}
\end{figure*}
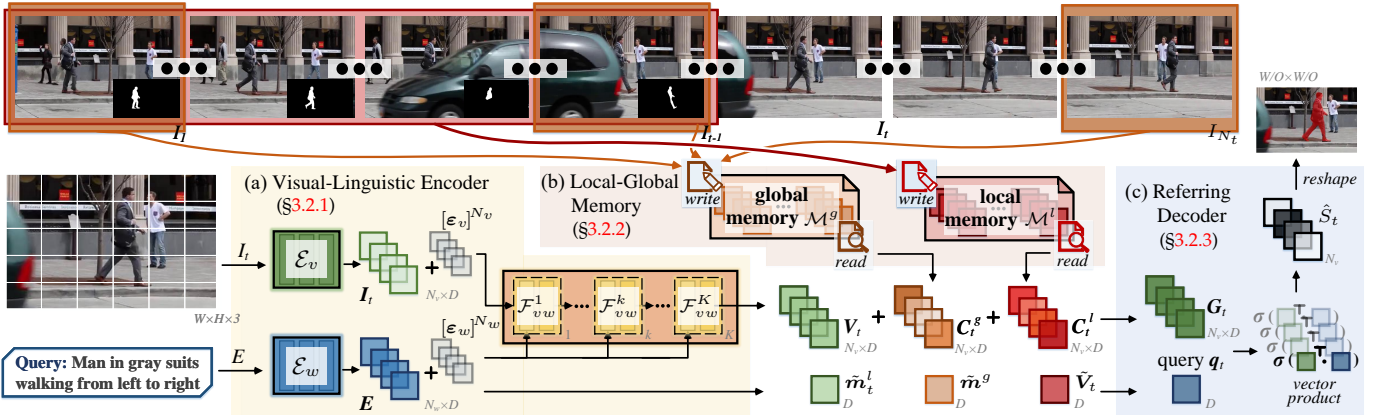

\subsection{Preliminary: Transformer Architecture}\label{sec:preliminary}

The core of Transformer~\cite{vaswani2017attention} is the attention mechanism. Given the query embedding $\Fbm^{q\!}\!\in_{\!}\!\RR^{N_{q\!} \x_{\!} D_{k\!}}$, key embedding $\Fbm^{k\!}\!\in_{\!}\!\RR^{N_{v\!} \x_{\!} D_{k\!}}$ and value embedding $\Fbm^{v\!}\!\in_{\!}\!\RR^{N_{v\!} \x_{\!} D_{v\!}}$,  the output of
a \textit{single-head attention} layer is computed as:
\begin{equation}\label{eq:attn}
    \Func{ATT}(\Fbm^{q\!}, \Fbm^{k\!}, \Fbm^{v\!})= \softmax{\left(\frac{\Fbm^{q\!}\Fbm^{k\top\!}}{\sqrt{\textit{D}_\textit{k}}}\right)}\Fbm^{v\!}.
\end{equation}
Combining several paralleled single-head attention, one can derive \textit{multi-head attention}. In Transformer, each encoder block has two layers, \ie, multi-head self-attention (MSA) and a multi-layer perceptron (MLP).  MSA is a variant of multi-head attention, where the query, key and value are from the same source. Each encoder block is formulated as:
\begin{equation}
    \begin{aligned}\label{eq:2}
        \Xbm' & = \text{LN}(\Xbm+\Func{MSA}(\Xbm)),   \\
        \Ybm  & = \text{LN}(\Xbm'+\Func{MLP}(\Xbm')),
    \end{aligned}
\end{equation}
where residual connection and layer-normalization are applied; $\Xbm$ and $\Ybm$ are the input and output, respectively.

The decoder block has a similar structure. A multi-head attention layer is first adopted, where a specific query embedding is generated to gather information from the encoder side, with the output of corresponding encoder block as key and value embeddings.

Despite the strong expressivity, applying Transformer~to LVS is not trivial. As the self-attention function (\cf~Eq.~\ref{eq:attn}) comes with quadratic time and space complexity, letting Transformer capture all the intra- and inter-modal relations is not feasible. To address the efficiency issue, some studies exploit sparse attention~\cite{child2019generating,beltagy2020longformer},  linear attention~\cite{katharopoulos2020transformers,wang2020linformer}, and recurrence~\cite{dai2019transformer,liang2022visual,wang2023lana}. With a similar spirit, we develop \textsc{Locater} that specifically focuses on LVS and models temporal and cross-modal dependencies with constant size memory and time complexity linear in sequence length.

\subsection{Local-Global Context Aware Transformer}\label{sec:Locater}

Given the input video $\{I_{t}\}_{t\!}$ and language expression~$E$,~our \textsc{Locater} mainly consists of three parts (see Fig.~\ref{fig:pipe}):~\textbf{i)} a \textit{visual-linguistic encoder} (\S\ref{sec:3.2.1}) that gradually fuses the linguistic embedding $\Ebm$ into visual embedding $\Ibm_t$ and genera- tes$_{\!}$ a$_{\!}$ language-enhanced$_{\!}$ visual$_{\!}$ feature$_{\!}$ $\Vbm_{t}$ for$_{\!}$ each$_{\!}$ frame$_{\!}$ $I_t$; \textbf{ii)} a \textit{local-global memory} (\S\ref{sec:3.2.2}) that gathers diverse temporal context$_{\!}$ from$_{\!}$ $\{I_{t}\}_{t}$ so$_{\!}$ as$_{\!}$ to$_{\!}$ render$_{\!}$ $\Vbm_{t}$ as$_{\!}$ a$_{\!}$ contextualized$_{\!}$ fea- ture$_{\!}$ $\Gbm_{t}$ and$_{\!}$ comprehend$_{\!}$ the$_{\!}$ expression$_{\!}$ $E$ into$_{\!}$ an$_{\!}$ expressive yet frame-specific query vector $\qbm_{t}$; and \textbf{iii)} a \textit{referring decoder} (\S\ref{sec:3.2.3}) that queries $\Gbm_{t\!}$ with $\qbm_{t\!}$ for mask prediction $\hat{S}_{t}$.

\subsubsection{Visual-Linguistic Encoder}\label{sec:3.2.1}
As illustrated in Fig.~\ref{fig:pipe} (a), the visual-linguistic encoder first extracts visual and linguistic features from each frame and the referring expression respectively, and then aggregates them into a compact, language-enhanced visual feature, through self-attention.

\noindent\textbf{Single-Modality Encoders.} For each frame $I_{t\!}\!\in\!\!\RR^{W_{\!} \x_{\!} H_{\!} \x_{\!} 3}$, a visual encoder $\Ecal_{v}$ is adopted to generate a raw visual embedding, \ie, $\Ibm_{t\!} \!=\! \Ecal_{v}(I_{t})\!\in_{\!}\!\RR^{N_{v\!}\x_{\!} D_{\!}}$. Here $\Ibm_{t\!}$ is a sequence of $D$-dimensional embeddings of $N_{v\!}$ image patches, \ie, $\Ibm_{t\!} \!=\!\{\Ibm_{t,p\!}\!\in_{\!}\!\RR^{D}\}_{p=1}^{N_{v\!}}$, where each patch is of $O_{\!}\x_{\!}O$ pixels and $N_v\!=\!WH/O^2$. At the same time, we forward the referring expression $E$ into a language encoder $\Ecal_{w}$ to get the corresponding linguistic embedding, \ie, $\Ebm \!=\! \Ecal_{{w}}(E)\!\in_{\!}\!\RR^{N_{w\!}\x_{\!} D}$.

\noindent\textbf{Cross-Modality Encoder.} To fuse heterogeneous features of each frame image and language expression early, we devise a cross-modality encoder $\Ecal_{vw}$. It adopts a Transformer encoder architecture, that captures all the correlations between image patches and language words, with several cascaded blocks. Specifically, the $k$-th module $\Fcal_{vw}^k$ is formulated as:
\begin{align}
        {\Fbm}_1^{k\!} & = \Fcal_{vw\_1}^{k\!}([\Fbm_2^{k-1\!} + [\bm{\varepsilon}_v]^{N_v}, \Ebm + [\bm{\varepsilon}_w]^{N_w}]) ~\in \RR^{N_v \x D},\label{eq:cross_1} \\
        {\Fbm}_2^{k\!} & = \Fcal_{vw\_2}^{k\!}({\Fbm}_1^k) ~\in \RR^{N_v \x D},\label{eq:cross_2}
\end{align}
where $\bm{\varepsilon}_v\!\in\!\RR^{D\!}$ and $\bm{\varepsilon}_w\!\in\!\RR^{D\!}$ are learnable token type vectors that indicate whether
an input token is from frame image or phrase.
$[\cdot,\cdot]$ indicates concatenation operation and $[\!\!~\cdot~\!\!]^{N\!}$ copies its input vector $N$ times before concatenation.
$\Fcal_{vw\_1\!}^k$ and $\Fcal_{vw\_2\!}^k$ are implemented as the Transformer encoder block (\textit{cf.}~\S\ref{sec:preliminary}), and ${\Fbm}_2^{0\!}\!=_{\!}\!\Ibm_t$. In $k$-th module, the first~block $\Fcal_{vw\_1}^k$ takes $(k_{\!}\!-_{\!}\!1)$-th module's output, $\Fbm_2^{k-1}$, and the text feature, $\Ebm$, as inputs, and conducts cross-modal fusion. Then it feeds its output, ${\Fbm}^k_1$, into the second block, $\Fcal_{vw\_2}^k$, for further refinement. Note that, for $\Fcal_{vw\_1}^k$, only the first $N_v$  output visual embeddings are preserved, while the last $N_w$ linguistic outputs are directly discarded.

With global operations, each module enriches every patch embedding with both the textual context and intra-frame context. By stacking $K$ attention-based modules, $\Ecal_{vw\!}$ repeats such visual feature enhancement process $K$ times, addressing the heterogeneity of visual and linguistic cues, and finally outputting a language-enhanced visual feature $\Vbm_{t}\!\in_{\!}\!\RR^{N_{v\!}\x_{\!} D\!}$ for each frame $I_t$.

\subsubsection{Local-Global Memory}\label{sec:3.2.2}
Although $\Vbm_t$ yields a powerful multi-modal representation, it is less suitable for LVS, as it is computed for each frame $I_t$ individually without accounting for temporal information. Directly using all the video patches $\{\Ibm_{t,p\!}\}_{t,p}$ as the visual tokens of the cross-modality attention encoder $\Ecal_{vw}$ will cause unaffordable computational and memory costs, since the complexity of self-attention computation scales quadratically with the sequence length (\textit{cf.}~\S\ref{sec:preliminary}). We thus frame a local-global memory module $\Mcal$, which helps \textsc{Locater} to capture the temporal context in an efficient manner.

Specifically, $\Mcal$ has two parts, \ie, $\Mcal^{l}$ and $\Mcal^{g}$ (Fig.~\ref{fig:pipe} (b)), for capturing local and global temporal context, respectively. Both the global memory $\Mcal^{g\!}\!=\!\{\mbm^g_n\!\in\!\RR^{D}\}_{n=1\!}^{N_g}$ and local memory $\Mcal^{l\!}\!=\!\{\mbm^l_n\!\in\!\RR^{D}\}_{n=1\!}^{N_l}$ have a fixed capacity, which is much smaller than the number of  patches in $\Ical$, \ie, $N_{g\!}\!\ll_{\!}\! N_vN_t$ and $N_{l\!}\!\ll_{\!}\! N_vN_t$. Each memory cell $\mbm^{g\!}$ ($\mbm^l$) is a highly summarized visual state from the whole video (several past frames), empowering \textsc{Locater} with a holistic understanding of video content. With learnable operations, the memory collects informative spatio-temporal context, dynamically interacts with the memory contents, and selectively determines which content should be preserved. This is in stark contrast to standard Transformer that stores immutable key and value vectors of all the tokens, causing huge storage size. \textsc{Locater} also uses the attention mechanism to retrieve its local-global memory, making it a fully attentional model.

\noindent\textbf{Global Memory.} To construct the global memory $\Mcal^{g}$, we first evenly sample a set of $N_{t'}$ frames from the whole video, which are supportive enough for comprehensive video content understanding, due to the redundancy among frames. Then we adopt a learnable \textit{write} controller to compress and store these sampled frames into $\Mcal^{g}$ as the global context. Specifically, given a sampled frame $I_{t'}$ with its language-enhanced visual feature $\Vbm_{t'\!}\!=_{\!}\!\{\Vbm_{t'\!,p}\!\in_{\!}\!\RR^{D}\}_{p=1}^{N_v}$, for each patch $p$, we update every memory cell in parallel:
\begin{equation}
    \begin{aligned}\label{eq:gw}
        \!\!\cbm^g_p \!   & =\! \Fcal_{c}^g(\Vbm_{t'\!,p}, [\mbm^g_n]_n)\!=\! \Wbm_c^g\big[\Vbm_{t'\!,p}, \Func{AP}([\mbm^g_n]_n)\big] ~\!\!\in \RR^{D}, \\
        \!\!\obm^g_p \!   & =\! \Fcal_{o}^g(\cbm^g_p, [\mbm^g_n]_n)\!=\! \big[\sigma (\cbm^g_p{}^{\top}\Wbm_o^g\mbm^g_n)\big]_n ~\!\!\in [0,1]^{N_g},\!\!      \\
        \!\!\mbm^g_n \! & \leftarrow\! \Fcal_{\text{AP}}([o^g_{n,p\!}\cdot\cbm^g_p+(1-o^g_{n,p})\cdot\mbm^g_n]_p) ~\in \RR^{D},
    \end{aligned}
\end{equation}
where $\Func{AP}$ refers to the average pooling operation, $\sigma(\cdot)$ is sigmoid activation, $[\cdot]_n$ indicates concatenating over variable $n$.
$\Wbm_c^{g\!}\!\in_{\!}\!\RR^{D\x2D}$ and $\Wbm_o^{g\!}\!\in_{\!}\!\RR^{D\x D}$ are learnable parameters, and $[\mbm^g_n]_n$ is initialized as learnable states.
Iteratively, $\Fcal_{c}^{g\!}$ creates a candidate $\cbm_p^{g\!}$ for memory update, based on $\Vbm_{t'\!,p\!}$ and $[\mbm^g_n]_n$; $\Fcal_{o}^{g\!}$ returns a gate vector $\obm_p^{g\!}\!=_{\!}\![o^g_{n,p}]_{n=1\!}^{N_g\!}$ according to the uniqueness of $\obm_p^{g\!}$ \wrt all~the memory slots $[\mbm^g_n]_n$; $o^g_{n,p}$ controls how much information to retain from prior state in $n$-{th} memory slot. In this way, the write controller learns to collect global context into $\Mcal^{g}$ and obviate redundant and noisy information.

After processing all the $N_{t'}$ sampled frames, the global memory $\Mcal^{g}$ is constructed and maintained unchanged during the whole segmentation process of $\Ical$. Thus \textsc{Locater} can pay continuous attention to global context, so as to better interpret long-term or complex activity related phrases (\eg, ``\textit{a boy is skiing  and suddenly falling}'' in Fig.$_{\!}$~\ref{fig:moti}) and handle video$_{\!}$ processing$_{\!}$ challenges$_{\!}$ (\eg,$_{\!}$ occlusion,$_{\!}$ fast$_{\!}$ motion,$_{\!}$ \etc).

\noindent\textbf{Local Memory.}
For the local memory $\Mcal^{l}$, it is constructed and updated on-the-fly. Specifically, given the last segmented frame $I_{t-1}$, with corresponding local memory state $\Mcal^{l}_{t-1\!}\!=_{\!}\!\{\mbm^l_{t-1,n\!}\!\in_{\!}\!\RR^{D}\}_{n=1}^{N_l}$, language-enhanced visual feature $\Vbm_{t-1\!}\!=_{\!}\!\{\Vbm_{t-1,p\!}\!\in_{\!}\!\RR^{D}\}_{p=1\!}^{N_v}$, and predicted segmentation mask $\hat{S}_{t-1\!}\!=\!\{s_{t-1,p}\!\in\![0,1]\}_{p=1}^{N_v}$ (\ie, a sequence of patch segments), the new memory state $\Mcal^{l}_{t}$ for current frame $I_{t}$ is obtained, again, by a learnable \textit{write} controller applied onto each memory state in parallel, like in Eq.~\ref{eq:gw}:
\begin{equation}
    \begin{aligned}\label{eq:lw}
        \cbm^l_p       & = \Fcal_{c}^l\big([\Vbm_{t-1,p}, \sbm_{t-1,p}],~~[\mbm^l_{t-1,n}]_n\big) ~\in \RR^{D}, \\
        \obm^l_p       & = \Fcal_{o}^l\big(\cbm^l_p,~~[\mbm^l_{t-1,n}]_n\big) ~\in [0,1]^{N_l},                   \\
        \mbm^l_{t,n} & = \Fcal_{\text{AP}([}o^l_{n,p\!}\cdot\cbm^l_p+(1-o^l_{n,p})\cdot\mbm^l_{t-1,n}]_p) ~\in \RR^{D},
    \end{aligned}
\end{equation}
where $\sbm_{t-1,p}\!\in\!\RR^{D}$ is a trainable segmentation mask embedding. It is obtained by applying a small FCN over $\hat{S}_{t-1\!}$. As a similar operation in Eq.~\ref{eq:gw}, the write controller learns to gather short-term context from past segmented frames and selectively update the contents of the local memory in a frame-by-frame manner.

With $\Mcal^{l}$, \textsc{Locater} can access and leverage local temporal context to comprehend simple action-related descriptions (\eg, ``\textit{a woman is running}''), and generate temporally coherent results with the aid of segmentation history.

\setlength{\intextsep}{10pt}
\begin{figure}[t]
    \begin{center}
        \includegraphics[width=1.\linewidth]{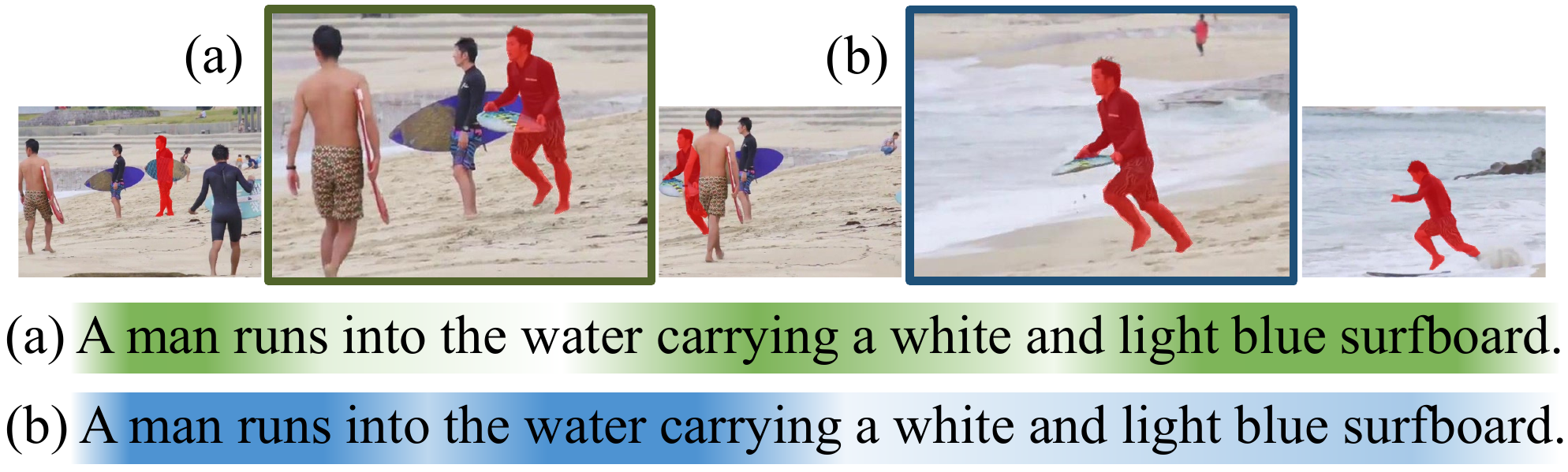}
    \end{center}
    \caption{\textbf{Attention visualization} {of frame-specific queries (Eq.~\ref{eq:query}).}}
    \label{fig:queryattn}
\end{figure}

\noindent\textbf{Contextualized Visual Embedding.} \textsc{Locater} next enriches the linguistic-enhanced visual feature $\Vbm_t$ of $I_t$ with local-global context, by looking up its current memory $\Mcal_t\!=\!\{\Mcal^{g}, \Mcal^{l}_{t}\}\!=\!\{\mbm_{t,n}\}_{n=1}^{N_g+N_l}$. An attention-based \textit{read} operator is used to retrieve context from the memory $\Mcal_t$:
\begin{equation}
    \begin{aligned}
        \cbm^g_t = \Func{ATT}(\Vbm_t,~[\mbm^g_{n}]_n,~[\mbm^g_{n}]_n) ~\in \RR^{N_v \x D}, \\
        \cbm^l_t = \Func{ATT}(\Vbm_t,~[\mbm^l_{t,n}]_n,~[\mbm^l_{t,n}]_n) ~\in \RR^{N_v \x D},
    \end{aligned}
\end{equation}
where $\cbm^g_t$ and $\cbm^l_t$ refer to the gathered global and local temporal context, respectively. Then we generate a contextualized visual embedding $\Gbm_t$ for frame $I_t$:
\begin{equation}
    \Gbm_t = \Vbm_t + \cbm^g_t + \cbm^l_t~\in \RR^{N_v \x D}.
\end{equation}
With this local-global memory-induced attention operation, $\Gbm_{t\!}$ encodes multi-modal information and rich temporal context, which is used in the decoder for  querying the referent.
\subsubsection{Referring Decoder}\label{sec:3.2.3}
\noindent\textbf{Contextualized Query Embedding.} The local-global~memory can not only enable contextualized visual embedding, but also help better parse the language expression. We first derive three compact vectors, \ie, $\tilde{\Vbm}_{t\!}\!\in_{\!}\!\RR^{1 \x D}$, $\tilde{\mbm}^{g\!}\!\in_{\!}\!\RR^{1 \x D}$ and $\tilde{\mbm}^l_{t}\in\RR^{1 \x D}$, from $\Vbm_t$, $\Mcal^{g}$ and $\Mcal^{l}_{t}$, respectively, to  summarize visual context in different temporal scales:
\begin{equation}
    \!\!\!\!\!\!\tilde{\Vbm}_{t\!} \!=\! \Func{AP}([\Vbm_{t,p}]_p),~\! \tilde{\mbm}^{g\!} \!=\! \Func{AP}([\mbm^g_{n}]_{n}),~\! \tilde{\mbm}^{l\!}_t \!=\! \Func{AP}([\mbm^l_{t,n}]_{n}),\!\!\!
\end{equation}
where $\Func{AP}$ indicates the average pooling operation, defined in Eq.~\ref{eq:gw}. Then, \textsc{Locater} interprets the expression $\Ebm_{\!}\!=_{\!}\!\{\Ebm_{w\!}\!\in_{\!}\!\RR^{1_{\!}\x_{\!}D}\}^{N_{w}}_{n=1\!}$ as a compact query vector $\qbm_t$, by using visual context guided attention to adaptively emphasize informative words:
\begin{equation}\label{eq:query}
    \begin{aligned}
        \!\!\!\!a_{t,w} \! & =\! \text{softmax}_w\big(([\tilde{\Vbm}_t, \tilde{\mbm}^g, \tilde{\mbm}^{l}_{t}]~\Wbm_1)(\Ebm_w\Wbm_2)^\top\big)\!\in\![0, 1],\!\!\!\!\\
        \!\!\qbm_t \!  & =\! \sum\nolimits_{n=1}^{N_w} a_{t,w} \cdot\Ebm_w \in\RR^{1 \x D},
    \end{aligned}
\end{equation}
where $\Wbm_{1\!}\!\in_{\!}\!\RR^{3D \x D\!}$ and $\Wbm_{2\!}\!\in_{\!}\!\RR^{D \x D\!}$ are learnable parameters, and $[a_{t,w}]_w$ encodes the importance of each word, by taking both temporal visual context, \ie, $\tilde{\mbm}^g$, $\tilde{\mbm}^{l}_{t}$, and specific content of frame $I_t$, \ie, $\tilde{\Vbm}_t$, into account.
As shown in Fig.~\ref{fig:queryattn}, with frame-specific attentions, the generated query vector $\qbm_t$ is expressive and particularly fits the corresponding frame $I_t$, instead of previous LVS methods~\cite{gavrilyuk2018actor,khoreva2018video,ningpolar,mcintosh2020visual} processing all the frames with a fixed query embedding.

\begin{table*}[t]
	\centering
	\caption{An example of applying semantic-role labeling to the video description for the construction of {A2D-S}$^+_{\text{S}}$ and {A2D-S}$^+_{~\!\text{T}}$. See \S\ref{sec:dataset_consturction} for details.}
	\resizebox{1.\textwidth}{!}{
		\setlength\tabcolsep{6pt}
		\renewcommand\arraystretch{1.4}
		\begin{tabular}{r|l|l}
			\hline\thickhline
			\multicolumn{3}{l}{\textbf{Original Description}: A boy is skiing on the snow and suddenly falling to the ground.} \\
			\hline
			\textbf{Verb }& \textbf{Semantic Role labeling} & \textbf{Semantic Groups} \\
			\hline
			is & A boy \textcolor{datared}{[Verb: is]} skiing on the snow and suddenly falling to the ground. & Filtered out \\
			skiing & \textcolor{datagreen}{[ARG0: A boy]} is \textcolor{datared}{[Verb: skiing]} \textcolor{orange}{[ARGM-LOC: on the snow]} and suddenly falling to the ground. & \textcolor{datagreen}{A boy} \textcolor{datared}{skiing} \textcolor{orange}{on the snow} \\
			falling & \textcolor{datagreen}{[ARG0: A boy]} is skiing on the snow and \textcolor{violet}{[ARGM-TMP: suddenly]} \textcolor{datared}{[Verb: falling]} \textcolor{datablue}{[ARG2: to the ground]}. & \textcolor{datagreen}{A boy} \textcolor{violet}{suddenly} \textcolor{datared}{falling} \textcolor{datablue}{to the ground} \\
			\hline
			\multicolumn{3}{l}{\textbf{Semantic-Role Labeled Description}: \textcolor{datagreen}{(boy, ARG0)}, \textcolor{datared}{(skiing, Verb)}, \textcolor{orange}{(on snow, ARGM-LOC)}, \textcolor{violet}{(suddenly, ARGM-TMP)}, \textcolor{datared}{(falling, Verb)}, \textcolor{datablue}{(to ground, ARG2)}} \\
			\hline
			\multicolumn{3}{l}{\textbf{Lemmatized Description}:  \textcolor{datagreen}{(boy, ARG0)}, \textcolor{datared}{(ski, Verb)}, \textcolor{orange}{(on snow, ARGM-LOC)}, \textcolor{violet}{(suddenly, ARGM-TMP)}, \textcolor{datared}{(fall, Verb)}, \textcolor{datablue}{(to ground, ARG2)}} \\
			\hline
		\end{tabular}
	}
	\label{tab:srl}
\end{table*}

\noindent\textbf{Referent Mask Decoding.} For mask generation, \textsc{Locater} queries the contextualized visual embedding $\Gbm_{t\!}\!=_{\!}\!\{\Gbm_{t,p\!}\!\in_{\!}\!\RR^{1\x D}\}_{p=1\!}^{N_v\!}$ with the frame-specific query vector $\qbm_{t\!}\!\in_{\!}\!\RR^{1\x D}$, by computing product distance between query-patch pairs:
\begin{equation}\label{eq:readout}
    ~~~~~~\hat{s}_{t,p} = \sigma(\Gbm_{t,p}\qbm_t^\top) ~~~\in [0,1].
\end{equation}
In this way, we can obtain a mask sequence $\hat{S}_{t\!}\!=_{\!}\!\{\hat{s}_{t,p}\}_{p=1\!\!}^{N_v\!}$ $\in_{\!}\![0,1]^{N_v\!}$ that collects all the patch-level responses. Recall that each patch is of $O\!\x\!O$ size and $N_{v\!}\!=_{\!}\!WH/O^{2\!}$. $\hat{S}_t$ is then reshaped into a 2D mask with $W/O\!\x\!H/O$ size and bilinearly upsampled to the original image resolution, \ie, $W\!\x\!H$, as the final segmentation result for the frame $I_t$.

\noindent\textbf{Deeply-Supervised Transformer Learning.} In practice, we find our Transformer based model is hard to train. Deeply-supervised learning~\cite{lee2015deeply,xie2015holistically}, \ie, introducing supervision signals into intermediate network layers, has been shown effective in CNN-based network training. Given the fact that Transformer also has a layer-by-layer architecture, we suppose that deeply-supervised learning can also help ease the learning of our model. To explore this idea, we separately forward the output features $\{\Fbm_{2}^k\}_{k=1}^{K}$ of all the $K$ modules in the cross-modality encoder $\Ecal_{vw}$ (Eq.~\ref{eq:cross_2}), into an auxiliary readout layer. The readout layer is a small MLP followed by sigmoid activation, and learns to predict a segmentation mask. For each training frame $I$, given the auxiliary outputs $\{\hat{S}_k\!\in\![0,1]^{W\x H\!}\}_{k=1}^{K}$ from $\Ecal_{vw}$ and final segmentation prediction $\hat{S}\!\in\![0,1]^{W\x H\!}$, the training loss is:
\begin{equation}\label{eq:loss}
    \Lcal = \sum\nolimits_I \big(\Lcal_{\text{CE}} (\hat{{S}}, {S}) + \lambda \sum\nolimits_{k} \Lcal_{\text{CE}} (\hat{{S}}_k, {S})\big),
\end{equation}
where $\Lcal$\sub{CE} refers to pixel-wise, binary cross-entropy loss, ${S}_{\!}\!\in_{\!}\!\{0,1\}^{H_{\!}\x_{\!}W\!\!}$ is the ground-truth mask, and $\lambda$ balances the two terms. Such a learning strategy can better supervise our early-stage cross-modal information fusion (\cf~\S\ref{sec:3.2.1}).

\subsection{Implementation Details}\label{sec:detail}
\noindent\textbf{Detailed Architecture.} Each video frame $I_t$ is resized to $320\!\x\!320$, and then patch-wise separated with a patch size of $16$, \ie, $O\!=\!16$ and $N_v\!=\!400$, as in~\cite{dosovitskiy2020image}. Each language description~is fixed to
$N_{w\!}\! =\! 20$ word length, with padding and truncation for the mismatched ones.
The visual encoder $\Ecal_{v}$ is implemented as six Transformer encoder blocks, while the linguistic encoder $\Ecal_{w}$ is implemented as a bi-LSTM~\cite{huang2015bidirectional} as in~\cite{ningpolar,wang2020context}. For the cross-modality encoder $\Ecal_{vw}$, it has $K_{\!}\!=_{\!}\!3$ modules. The hidden sizes of all the modules are set to $D\!=\!768$. For the global memory $\Mcal^g$ and local memory $\Mcal^{l\!}$, we set the capacity as ${N_{g\!}\!=_{\!}\!1.5N_v}$ and ${N_{l\!}\!=_{\!}\!2N_v}$ respectively. Unless otherwise specified, we sample representative frames with an interval of \cnum{10} frames to construct $\Mcal^g$
, \ie, $N_{t'}\approx N_t/10$. Related experiments can be found in \S\ref{sec:ablation}.

\noindent\textbf{Training Details.}
Our model is trained for $30$ epochs using Adam optimizer~\cite{kingma2014adam} with initial learning rate \cnum{4e-5}, batch size \cnum{32} and weight decay \cnum{1e-4}. We adopt polynomial annealing policy~\cite{liang2022gmmseg} to schedule the learning rate. The $\lambda$ in Eq.~\ref{eq:loss} is set to \cnum{0.4} by default. Random horizontal flipping is employed as data augmentation where video frames are flipped with the corresponding positional description being modified from ``\textit{right}" to ``\textit{left}" and vice versa.

\noindent\textbf{Inference.}
During inference, for each video, \textsc{Locater} first builds the global memory $\Mcal^g$ and maintains it unchanged. Then, \textsc{Locater} conducts segmentation in a sequential manner. The local memory $\Mcal^l$ is gradually updated during segmentation. The continuous segmentation prediction mask $\hat{S}$ for each frame $I$ is binarized with a threshold of $0.5$.

\section{Our A2D-S$^+$ Dataset}\label{sec:a2d+}

Till now, A2D-S is the most popular LVS dataset. However, in practice, we find that the majority of its \texttt{test} videos (\cnum{459} of \cnum{757}) only  contain one \textit{single} actor. With these trivial cases, such cross-modal task tends to degrade to a single-modal problem of novel object segmentation. Moreover, many of the remaining videos only contain very few objects yet with distinctive semantics. To better examine the visual grounding capability of LVS models, we construct a \textit{harder} dataset -- A2D-S$^+$. It consists of three subsets, \ie, {A2D-S}$^+_{\text{M}}$, {A2D-S}$^+_{\text{S}}$, and {A2D-S}$^+_{~\!\text{T}}$, which are all built upon A2D-S but fully aware of the limitation of A2D-S. Specifically, each video in {A2D-S}$^{+\!}$ is selected/created to contain multiple instances of the same object or action category. Thus our {A2D-S}$^{+\!}$ places a higher demand for the grounding ability of LVS models as the distinction among the similar instances is necessary. Next, we will detail the construction process in \S\ref{sec:dataset_consturction} and discuss dataset features and statistics in \S\ref{sec:dataset_statistics}.

\begin{figure*}
    \begin{center}
        \includegraphics[width=0.98\linewidth]{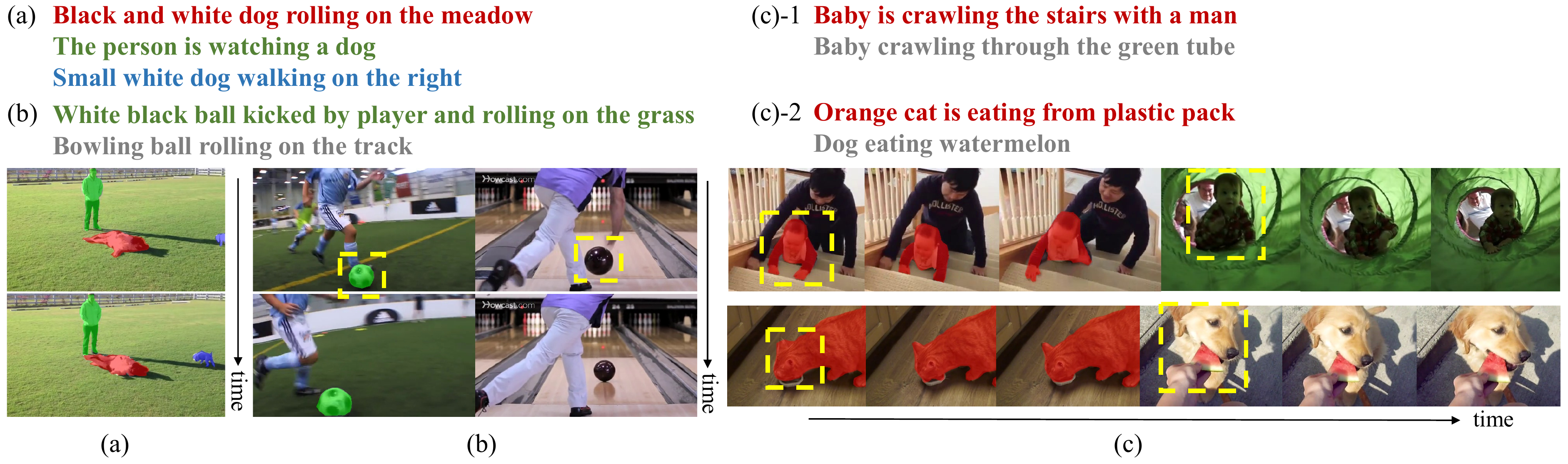}
    \end{center}
    \caption{\textbf{Representative video examples} {from our proposed (a) A2D-S$^+_{\text{M}}$, (b) A2D-S$^+_{\text{S}}$, and (c) A2D-S$^+_{~\!\text{T}}$ datasets (\S\ref{sec:dataset_consturction}).}}
    \label{fig:a2dexample}
\end{figure*}

\subsection{Dataset Construction}\label{sec:dataset_consturction}

We first$_{\!}$ curate$_{\!}$ videos$_{\!}$ with$_{\!}$ multiple$_{\!}$ actors$_{\!}$ from$_{\!}$ A2D-S$_{\!}$ \texttt{test}$_{\!\!}$ as the subset {A2D-S}$^{+\!}_{\text{M}\!\!\!}$~({A2D-S}$^+_{\text{Multiple}}$). Although each video~in {A2D-S}$^{+\!}_{\text{M}\!\!\!}$ has multiple instances, most of these instances~are with different semantic categories, making the distinction among them relatively easy. To mitigate this, we further create two challenging subsets: {A2D-S}$^+_{\text{S}}$ and {A2D-S}$^+_{~\!\text{T}}$, by~emp- loying a contrasting sampling strategy~\cite{sadhu2020video}, which synthesizes new test samples from videos that contain similar~but not exactly the same instances as described by the language expression. Specifically, the contrasting sampling based construction process of {A2D-S}$^+_{\text{S}}$ and {A2D-S}$^+_{~\!\text{T}}$ includes three steps: \textbf{i)} parse free-form language descriptions into semantic-roles (\eg, \textit{who} did \textit{what} to \textit{whom}); \textbf{ii)} for each query video, sample a video with the description of the same semantic-roles structure as the queried description, but the role is realized by a different noun or a verb; and \textbf{iii)} generate a new sample by concatenating the query and sampled videos along the width axis (for {A2D-S}$^+_{\text{S}}$) or time axis (for {A2D-S}$^+_{~\!\text{T}}$).

\noindent\textbf{Semantic-Role Labeling (SRL).} We first parse each free- form$_{\!}$ language$_{\!}$ description$_{\!}$ in$_{\!}$ A2D-S$_{\!}$ \texttt{test}$_{\!}$  into$_{\!}$ semantic-roles$_{\!}$~\cite{strubell2018linguistically}, \ie, given the description, answer the high-level question of ``who ({\color{datagreen}{\texttt{Arg0}}}) did what ({\color{datared}{\texttt{Verb}}}) to whom ({\color{datablue}{\texttt{Arg1}}}) at where ({\textcolor{orange}{\texttt{ARGM}-\texttt{LOC}}}) and when ({\textcolor{violet}{\texttt{ARGM-TMP}}})\footnote{There are other types of semantic-roles that are not considered in our case, \eg, How (\texttt{ARGM}-\texttt{MNR}), Why (\texttt{ARGM}-\texttt{CAU}), which only applicable for {\app}$1\%$ of the language queries in  A2D-S \texttt{test}.}''. In particu- lar, we first use a BERT-based~\cite{devlin2018bert} semantic-role labeling model~\cite{shi2019simple} to predict the semantic-roles for each language description. The model is trained on OntoNote5~\cite{pradhan2013towards} with PropBank annotations~\cite{palmer2005proposition}. Then the labeled sentences~are further cleaned by: \textbf{i)} removing words without any roles, \eg, ``is'', ``are'', ``a", ``the"; \textbf{ii)} abandoning non-semantic-role labeled sentences; and \textbf{iii)} lemmatization~\cite{qi2020stanza}. An example of our semantic-role labeling process is given in Table~\ref{tab:srl}.

\noindent\textbf{Contrasting Sampling.} After assigning semantic-roles to language descriptions of A2D-S$_{\!}$ \texttt{test}$_{\!}$ videos, we conduct contrasting sampling to find similar but not exactly the same examples, according to their parsed semantic-role structures.  Specifically, for each description, we sample one other description from the dataset that contains at least $_{\!}$one, but not all, of semantic roles, which are realized with the same phrase.
For example, with a description \{(boy, {\color{datagreen}{\texttt{Arg0}}}), (run, {\color{datared}{\texttt{Verb}}}), (\textbf{on grass}, {\textcolor{orange}{\texttt{ARGM-LOC}}})\}, an eligible sampled description can be \{(boy, {\color{datagreen}{\texttt{Arg0}}}), (run, {\color{datared}{\texttt{Verb}}}), (\textbf{on snow}, {\textcolor{orange}{\texttt{ARGM-LOC}}})\}.
Finally, all sampled example pairs are checked by human for diminishing visual and linguistic confusions.

\noindent\textbf{Concatenation Strategies.} For each valid contrasting example pair, two concatenation strategies are adopted to merge them into one video, \ie, concatenating along the space axis as an instance of A2D-S$^+_{\text{S}}$ (A2D-S$^+_{\text{Spatial}}$), and along the time axis as an instance of A2D-S$^+_{\!~\text{T}}$ (A2D-S$^+_{\!~\text{Temporal}}$). Specifically:

\begin{itemize}[leftmargin=*]
	\setlength{\itemsep}{0pt}
	\setlength{\parsep}{1pt}
	\setlength{\parskip}{1pt}
	\setlength{\leftmargin}{-10pt}
    \item For {A2D-S$^+_{\text{S}}$}, videos are concatenated along the width axis exclusively, since height-axis concatenation might violate the natural up-down order in real world, \eg, \textit{sky} should be on top of \textit{grass}. The sampled videos are resized to have the same height as the query videos, and truncated or padded to fit the time durations of the query videos.
    \item For {A2D-S$^+_{\!~\text{T}}$}, videos are concatenated along the time axis. For a query video with $N_t$ frames, the sampled contrasting video is resized into the same resolution as the query video. There will be $2N_t$ frames in each newly constructed example, in which $N_t$ frames are from the query video, and the other $N_t$ frames are sampled from the contrasting video. For the frames from the contrasting videos, LVS models are expected to predict all-zero masks to indicate that there is no target referent.
\end{itemize}

Notably, there is no constraint on the concatenation order of videos. By default, we put the query video at the left and ahead, for spatial and temporal concatenation, respectively. We provide representative examples for A2D-S$^+_{\text{M,S,T}}$ in~Fig.~\ref{fig:a2dexample}.

\begin{table}[t]
	\centering
	\caption{\textbf{Dataset statistics} of A2D-S \texttt{test} and our A2D-S$^+_{\text{M,S,T}}$ (\S\ref{sec:dataset_statistics}).}
	\resizebox{0.48\textwidth}{!}{
		\setlength\tabcolsep{6pt}
		\renewcommand\arraystretch{1.1}
		\begin{tabular}{|l||c|ccc|}
			\hline\thickhline
			\rowcolor{mygray}
			\multicolumn{1}{|c||}{\textbf{Dataset}} & {A2D-S} \texttt{test}& A2D-S$^+_{\text{M}}$ & A2D-S$^+_{\text{S}}$ & A2D-S$^+_{\!~\text{T}}$\\
			\hline
			\hline
			$\#$Videos & 737 & 257 & 1,290 & 1,293 \\
			$\#$Queries & 1,295 & 856 & 1,295 & 1,295 \\
			$\#$Objects/vid & 1.76 & 3.33 & 5.02 & 5.13 \\
			\hline
		\end{tabular}
	}
	\label{tab:statistic}
\end{table}

\newcommand{\pub}[1]{{\color{gray}{\tiny{[{#1}]}}}}

\begin{table*}[t]
    \centering
    \caption{{Quantitative results} on A2D-S \texttt{test}~\cite{gavrilyuk2018actor} (\S\ref{sec:a2d}). (*: ViT-B with first 6 layers as backbone. Same for other tables.)}
    \label{tab:a2dsota}
    \tablestyle{0.98}{4}{1.1}{
        \begin{tabular}{|rl||c|ccccc|c|cc|c|}
            \hline\thickhline
            \rowcolor{mygray}
                                        &                           & & \multicolumn{5}{c|}{\textbf{Overlap}} & \textbf{mAP}  & \multicolumn{2}{c|}{\textbf{IoU}} & \\
            \rowcolor{mygray}
            \multicolumn{2}{|c||}{\multirow{-2}{*}{\textbf{Method}}}  & \multirow{-2}{*}{\textbf{Backbone}} & P@0.5 & P@0.6 & P@0.7 & P@0.8 & P@0.9 & 0.5:0.95 & Overall & Mean & \multirow{-2}{*}{\textbf{FPS}} \\
            \hline \hline
            $^{\dag}$Hu \etal~\cite{hu2016segmentation}        \!\!\!&\!\!\pub{ECCV16}    & VGG-16~\cite{simonyan2014very} & 34.8 & 23.6 & 13.3 & 3.3  & 0.1  & 13.2 & 47.4 & 35.0 & - \\
            $^{\dag}$Li \etal~\cite{li2017tracking}            \!\!\!&\!\!\pub{CVPR17}    & VGG-16~\cite{simonyan2014very} & 38.7 & 29.0 & 17.5 & 6.6  & 0.1  & 16.3 & 51.5 & 35.4 & - \\
            Gavrilyuk \etal~\cite{gavrilyuk2018actor} \!\!\!&\!\!\pub{CVPR18}    & I3D~\cite{carreira2017quo} & 50.0 & 37.6 & 23.1 & 9.4  & 0.4  & 21.5 & 55.1 & 42.6 & - \\
            Wang \etal~\cite{wang2019asymmetric}      \!\!\!&\!\!\pub{ICCV19}    & I3D~\cite{carreira2017quo} & 55.7 & 45.9 & 31.9 & 16.0 & 2.0  & 27.4 & 60.1 & 49.0 & 8.64 \\
            Wang \etal~\cite{wang2020context}         \!\!\!&\!\!\pub{AAAI20}    & I3D~\cite{carreira2017quo} & 60.7 & 52.5 & 40.5 & 23.5 & 4.5  & 33.3 & 62.3 & 53.1 & 7.18 \\
            McIntosh \etal~\cite{mcintosh2020visual}  \!\!\!&\!\!\pub{CVPR20}    & I3D~\cite{carreira2017quo} & 52.6 & 45.0 & 34.5 & 20.7 & 3.6  & 30.3 & 56.8 & 46.0 & - \\
            Ning \etal~\cite{ningpolar}               \!\!\!&\!\!\pub{IJCAI20}   & I3D~\cite{carreira2017quo} & 63.4 & 57.9 & 48.3 & 32.2 & 8.3  & 38.8 & 66.1 & 52.9 & 5.42 \\
            Hui \etal~\cite{hui2021collaborative}     \!\!\!&\!\!\pub{CVPR21}    & I3D~\cite{carreira2017quo} & 65.4 & 58.9 & 49.7 & 33.3 & 9.1  & 39.9 & 66.2 & 56.1 & 8.07 \\
            Yang \etal~\cite{yang2021hierarchical}    \!\!\!&\!\!\pub{BMVC21}    & ResNet-101~\cite{he2016deep} & 61.1 & 55.9 & 48.6 & 34.2 & 12.0 & -    & 67.9 & 52.9 & - \\
            \multicolumn{2}{|c||}{\textbf{\textsc{Locater}}} & $^{*}$ViT-B~\cite{dosovitskiy2020image} & \textbf{70.9} & \textbf{64.0} & \textbf{52.5} & \textbf{35.1} & \textbf{10.1} & \textbf{42.8} & \textbf{69.0} & \textbf{59.7} & \textbf{10.1} \\
            \hline
            MTTR~\cite{botach2022end} \!\!\!&\!\!\pub{CVPR22} & Video-Swin-T~\cite{liu2022video} & 75.4 & 71.2 & 63.8 & 48.5 & 16.9 & 46.1 & 72.0 & 64.0 & 6.69 \\
            ReferFormer~\cite{wu2022language} \!\!\!&\!\!\pub{CVPR22} & Video-Swin-T~\cite{liu2022video} & 76.0 & 72.2 & 65.4 & \textbf{49.8} & \textbf{17.9} & 48.6 & \textbf{72.3} & 64.1 & 6.54 \\
            \multicolumn{2}{|c||}{\textbf{\textsc{Locater}}} & Video-Swin-T~\cite{liu2022video} & \textbf{76.5} & \textbf{72.8} & \textbf{66.3} & 49.5 & 17.1 & \textbf{48.8} & 71.6 & \textbf{64.2} & 7.92 \\
            \hline
        \end{tabular}
    }
\end{table*}

\subsection{Dataset Features and Statistics}\label{sec:dataset_statistics}
\noindent\textbf{Dataset Features.} Through the above contrasting sampling process, A2D-S$^{+\!}$ has two distinctive characteristics:
\begin{itemize}[leftmargin=*]
	\setlength{\itemsep}{0pt}
	\setlength{\parsep}{1pt}
	\setlength{\parskip}{1pt}
	\setlength{\leftmargin}{-10pt}
    \item \textit{{Dense Grounding-ability Required:}}
A2D-S${^+\!}$ is manufactured to guarantee that each video does contain multiple semantically similar objects where grounding among them is necessarily required.

    \item \textit{{Low Human-labour Cost:}}
    The entire dataset creation process is semi-automatic so that the human labour cost is greatly reduced. Only a few inevitable efforts have been made to diminish the visual and linguistic ambiguity.
\end{itemize}
\noindent\textbf{Dataset Statistics.}
The detailed statistics of $\text{A2D-S}^+_\text{M,S,T}$ are shown in Table~\ref{tab:statistic}. As our  A2D-S${^+\!}$ is only used to evaluate LVS models, the statistic of A2D-S \texttt{test} is also provided for clear comparison. As seen, compared to A2D-S \texttt{test}, the newly constructed $\text{A2D-S}^+_\text{M,S,T}$ significantly increase the number of existing semantically similar objects in each video, \ie, \cnum{1.76} \textit{vs} \cnum{3.33}/\cnum{5.02}/\cnum{5.13}, which {together} provide a stronger testing bed for evaluating LVS methods.

\section{Experiment}
\noindent\textbf{Overview.} To thoroughly examine the efficacy of \textsc{Locater}, we first report quantitative results on three standard LVS datasets, \ie, A2D Sentences (A2D-S)~\cite{gavrilyuk2018actor} (\S\ref{sec:a2d}), J-HMDB Sentences (J-HMDB-S)~\cite{gavrilyuk2018actor} (\S\ref{sec:jhmdb}), and Refer-Youtube-VOS (R-YTVOS)~\cite{seo2020urvos} (\S\ref{sec:urvos}), followed by qualitative results (\S\ref{sec:qua_vis}).
And we further perform experiments on our proposed A2D-S$^+$ dataset in \S\ref{sec:contra_setting}.
Then in \S\ref{sec:challenge}, we report the model performance on the RVOS Track in YTB-VOS\sub{21} Challenge$_{\!}$~\cite{vosc2021}, where our \textsc{Locater} based solution achieved the 1\textit{st} place.
Later, in \S\ref{sec:ablation}, we conduct a set of ablative studies to examine the core ideas and essential components of \textsc{Locater}.
We finally analyse several typical failure modes in \S\ref{sec:discussion}.

\begin{table*}[t]
    \centering
    \caption{{Quantitative results} on J-HMDB-S~\cite{gavrilyuk2018actor} (\S\ref{sec:jhmdb}). }
    \label{tab:jhmdbsota}
    \tablestyle{0.98}{5}{1.1}{
        \begin{tabular}{|rl||c|ccccc|c|cc|}
            \hline\thickhline
            \rowcolor{mygray}
                                        &                           & & \multicolumn{5}{c|}{\textbf{Overlap}} & \textbf{mAP}  & \multicolumn{2}{c|}{\textbf{IoU}} \\
            \rowcolor{mygray}
            \multicolumn{2}{|c||}{\multirow{-2}{*}{\textbf{Method}}}  & \multirow{-2}{*}{\textbf{Backbone}} & P@0.5 & P@0.6 & P@0.7 & P@0.8 & P@0.9 & 0.5:0.95 & Overall & Mean \\
            \hline \hline
            $^{\dag}$Hu \etal~\cite{hu2016segmentation}        \!\!\!\!&\!\!\!\pub{ECCV16}    & VGG-16~\cite{simonyan2014very} & 63.3 & 35.0 & 8.5  & 0.2  & 0.0 & 17.8 & 54.6 & 52.8 \\
            $^{\dag}$Li \etal~\cite{li2017tracking}            \!\!\!\!&\!\!\!\pub{CVPR17}    & VGG-16~\cite{simonyan2014very} & 57.8 & 33.5 & 10.3 & 0.6  & 0.0 & 17.3 & 52.9 & 49.1 \\
            Gavrilyuk \etal~\cite{gavrilyuk2018actor} \!\!\!\!&\!\!\!\pub{CVPR18}    & I3D~\cite{carreira2017quo} & 71.2 & 51.8 & 26.4 & 3.0  & 0.0 & 26.7 & 55.5 & 57.0 \\
            Wang \etal~\cite{wang2019asymmetric}      \!\!\!\!&\!\!\!\pub{ICCV19}    & I3D~\cite{carreira2017quo} & 75.6 & 56.4 & 28.7 & 3.4  & 0.0 & 28.9 & 57.6 & 58.4 \\
            Wang \etal~\cite{wang2020context}         \!\!\!\!&\!\!\!\pub{AAAI20}    & I3D~\cite{carreira2017quo} & 74.2 & 58.7 & 31.6 & 4.7  & 0.0 & 30.1 & 55.4 & 57.6 \\
            McIntosh \etal~\cite{mcintosh2020visual}  \!\!\!\!&\!\!\!\pub{CVPR20}    & I3D~\cite{carreira2017quo} & 67.7 & 51.3 & 28.3 & 5.1  & 0.0 & 26.1 & 53.5 & 55.0 \\
            Ning \etal~\cite{ningpolar}               \!\!\!\!&\!\!\!\pub{IJCAI20}   & I3D~\cite{carreira2017quo} & 69.1 & 57.2 & 31.9 & 6.0  & 0.1 & 29.4 & -    & -    \\
            Hui \etal~\cite{hui2021collaborative}     \!\!\!\!&\!\!\!\pub{CVPR21}    & I3D~\cite{carreira2017quo} & 78.3 & 63.9 & 37.8 & 7.6  & 0.0 & 33.5 & 59.8 & 60.4 \\
            Yang \etal~\cite{yang2021hierarchical}    \!\!\!\!&\!\!\!\pub{BMVC21}    & ResNet-101~\cite{he2016deep} & 81.9 & 73.6 & \textbf{54.2} & \textbf{16.8} & \textbf{0.4} & -    & 65.2 & 62.7 \\
            \multicolumn{2}{|c||}{\textbf{\textsc{Locater}}} & $^{*}$ViT-B~\cite{dosovitskiy2020image} & \textbf{89.3} & \textbf{77.2} & {50.8} & {10.6} & {0.2} & \textbf{36.3} & \textbf{67.3} & \textbf{66.3} \\
            \hline
            MTTR~\cite{botach2022end} \!\!\!\!&\!\!\!\pub{CVPR22} & Video-Swin-T~\cite{liu2022video} & \textbf{93.9} & 85.2 & 61.6 & 16.6 & 0.1 & 39.2 & 70.1 & \textbf{69.8} \\
            ReferFormer~\cite{wu2022language} \!\!\!\!&\!\!\!\pub{CVPR22} & Video-Swin-T~\cite{liu2022video} & 93.3 & 84.2 & 61.4 & 16.4 & 0.3 & 39.1 & 70.0 & 69.3 \\
            \multicolumn{2}{|c||}{\textbf{\textsc{Locater}}} & Video-Swin-T~\cite{liu2022video} & 93.6 & \textbf{85.9} & \textbf{61.9} & \textbf{16.8} & \textbf{0.3} & \textbf{39.4} & \textbf{70.8} & 69.6 \\
            \hline
        \end{tabular}
    }
\end{table*}

\noindent\textbf{Evaluation Criteria.} For A2D-S and J-HMDB-S, we follow conventions~\cite{gavrilyuk2018actor,wang2020context,ningpolar,wang2019asymmetric} to use intersection-over-union (IoU) and precision for evaluation.  We report \textit{overall IoU} as the ratio of the total intersection area divided by the total union area over testing samples, as well as \textit{mean IoU} as the average IoU of all samples. We also measure precision@$K$ as the percentage of testing samples whose IoU scores are higher than an overlap threshold $K$. We report precision at five thresholds ranging from \cnum{0.5} to \cnum{0.9} and mean average precision (mAP) over $0.50$:~$0.05$:~$0.95$. For R-YTVOS and YTB-VOS\sub{21}, we follow their standard evaluation protocols to report the region similarity ($\Jcal$), contour accuracy ($\Fcal$) and their average score $\Jcal\&\Fcal$ over all video sequences~\cite{perazzi2016benchmark}.

\subsection{Results on A2D-S Dataset}\label{sec:a2d}
We first conduct experiments on A2D-S~\cite{gavrilyuk2018actor}, which is the most popular dataset in the field of LVS.

\noindent\textbf{Dataset.}
A2D-S contains \cnum{3782} videos with \cnum{8} action classes performed by \cnum{7} actors, and \cnum{6655} actor and action related descriptions. In each video,  \cnum{5} to \cnum{7} frames are provided with segmentation annotations. As in~\cite{wang2019asymmetric}, we use \cnum{3017}/\cnum{737} split for \texttt{train}/\texttt{test}, and ignore the \cnum{28} unlabeled videos.

\newcommand{\tpm}[1]{{\,$\pm$\,$_{\!}$#1}}

\noindent\textbf{Quantitative Performance.}
Table~\ref{tab:a2dsota} reports the comparison results of \textsc{Locater} against 3D CNN based methods and two latest fully attentional works on the A2D-S \texttt{test}.
For fair comparisons with the concurrent competitors~\cite{wu2022language,botach2022end}, we report the model performance with the strong Video-Swin-T~\cite{liu2022video} backbone.
As seen, \textsc{Locater} yields state-of-the-art performance for all metrics when compared with 3D CNN based methods. Concretely, it advances the SOTA in mAP by \bpo{2.9}, Mean IoU by \bpo{3.6}, Overall IoU by \bpo{2.8}, and also produces great improvements in terms of precision scores under all overlap thresholds. Equipped with the strong Video-Swin-T backbone, \textsc{Locater} achieves comparable or even better performance than the contemporary Transformer based methods~\cite{botach2022end,wu2022language}.
For completeness, we report here the standard deviations of the best performed variants (Row 13): \tpm{0.354} and \tpm{0.518} in terms of mIoU and oIoU.
These experimental results well demonstrate the superiority of \textsc{Locater} on local-global semantics understanding brought by the memory design.

\noindent\textbf{Runtime Analysis.}
Table~\ref{tab:a2dsota} reports the efficiency comparison with several famous methods, including four FCN-style models~\cite{wang2019asymmetric,wang2020context,ningpolar}, and two latest fully attentional models~\cite{wu2022language,botach2022end}. For fairness, we conduct runtime analysis on a video clip of \cnum{36} frames with resolution $512\!\x\!512$ and a text sequence of \cnum{20} words length. The window size of 3D CNN/Transformer backbones, \ie, I3D~\cite{carreira2017quo} and Video-Swin-T~\cite{liu2022video}, are all set to \cnum{16}. The inference speed is measured on a single NVIDIA GeForce RTX 2080 Ti GPU.
As seen, \textsc{Locater} is much faster than existing LVS methods, owing to its memory-augmented fully attentional architecture design.

\subsection{Results on J-HMDB-S Dataset}\label{sec:jhmdb}
To investigate the generalization ability of \textsc{Locater}, we also report the performance of our A2D-S trained model on J-HMDB-S~\cite{gavrilyuk2018actor}, following~\cite{ningpolar,gavrilyuk2018actor,wang2020context,mcintosh2020visual}.

\noindent\textbf{Dataset.} J-HMDB-S contains $928$ short videos with $21$ different action categories and $928$ language descriptions.

\noindent\textbf{Quantitative Performance.}
Table~\ref{tab:jhmdbsota} presents performance comparison on J-HMDB-S. All models are trained under the same setting on A2D-S \texttt{train} exclusively without fine-tuning. As seen, our model surpasses other competitors across most metrics. Notably, \textsc{Locater} yields Mean IoU \bpo{66.3}, Overall IoU \bpo{67.3} and mAP \bpo{36.3}, while the corresponding scores for the previous SOTA method~\cite{wu2022language} are \po{62.7}, \po{65.2} and \po{33.5}, respectively. Compared with the recent works~\cite{wu2022language,botach2022end}, competitive performance is also achieved. These results confirm again the effectiveness of our \textsc{Locater}.

\begin{table}[t]
    \centering
    \caption{\textbf{Quantitative results} {on R-YTVOS \texttt{val}~\cite{seo2020urvos} (\S\ref{sec:urvos}).} }
    \hspace{-1.5em}
    \tablestyle{0.5}{2}{1.1}{
        \begin{tabular}{|rl|c||c|cc|}
            \hline\thickhline
            \rowcolor{mygray}
            \multicolumn{2}{|c|}{\textbf{Method}}                      & Backbone          & $\Jcal\&\Fcal$ & $\Jcal$ & $\Fcal$ \\
            \hline \hline
            Seo \etal~\cite{seo2020urvos} \!\!&\!\!\pub{ECCV20}         & ResNet-50~\cite{he2016deep} & 48.9           & 47.0    & 50.8    \\
            Hui \etal~\cite{hui2021collaborative} \!\!&\!\!\pub{CVPR21} & I3D~\cite{carreira2017quo} & 36.7           & 35.0    & 38.5    \\
            Li \etal~\cite{li2022you} \!\!&\!\!\pub{AAAI22}             & ResNet-50~\cite{he2016deep} & 48.6           & 47.5    & 49.7    \\
            \textbf{\textsc{Locater}}                                  && $^{*}$ViT-B~\cite{dosovitskiy2020image} & \textbf{50.0}  & \textbf{48.8} & \textbf{51.1}  \\
            \hline
            MTTR~\cite{botach2022end} \!\!&\!\!\pub{CVPR22}             & Video-Swin-T~\cite{liu2022video} & 55.3 & 54.0 & 56.6    \\
            ReferFormer~\cite{wu2022language} \!\!&\!\!\pub{CVPR22}     & Video-Swin-T~\cite{liu2022video} & 56.0 & 54.8 & 57.3     \\
            \textbf{\textsc{Locater}}                        && Video-Swin-T~\cite{liu2022video} & \textbf{56.5} & \textbf{54.8} & \textbf{58.1} \\
            \hline
        \end{tabular}
    }
    \label{tab:youtubesota}
\end{table}

\subsection{Results on R-YTVOS Dataset}\label{sec:urvos}
We further train and test our model on a recently new proposed large-scale dataset, R-YTVOS~\cite{seo2020urvos}.

\noindent\textbf{Dataset.} R-YTVOS has \cnum{3978} videos, with 131K segmentation masks and \cnum{15}K expressions. Only \cnum{3471}/\cnum{202} \texttt{train}/ \texttt{val} videos are public available for training and evaluation.

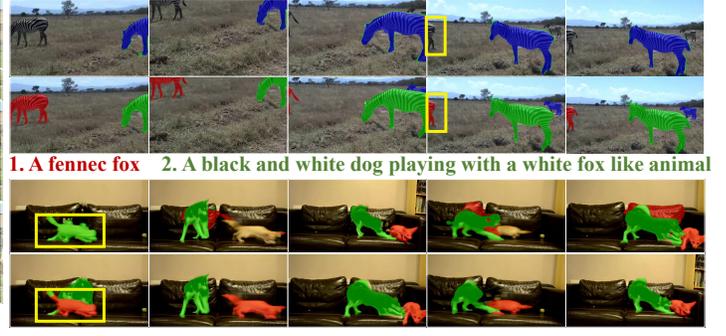
\begin{figure*}[t]
    \begin{center}
        \includegraphics[width=1\linewidth]{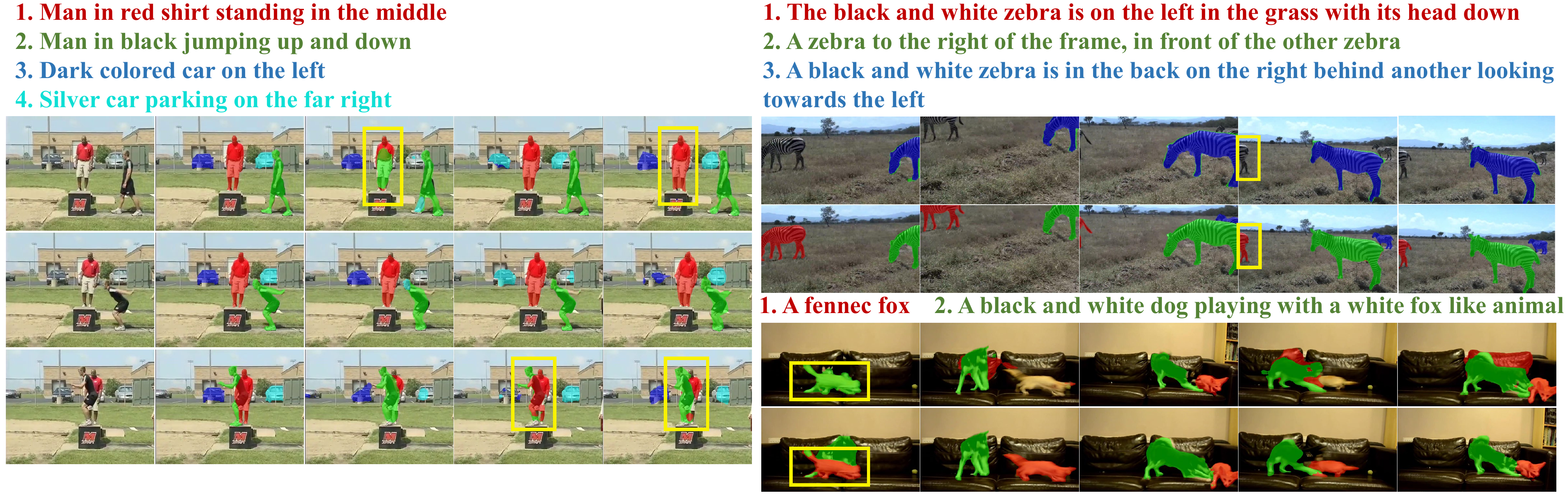}
        \put(-387, 2){\scriptsize~\cite{wang2019asymmetric}}
        \put(-338, 2){\scriptsize~\cite{hui2021collaborative}}
        \put(-515, 2){\scalebox{.70}{{Input Sequence}}}
        \put(-464, 2){\scalebox{.70}{{Ground Truth}}}
        \put(-407, 2){\scalebox{.70}{{ACGA}}}
        \put(-357, 2){\scalebox{.70}{{CSTM}}}
        \put(-307, 2){\scalebox{.70}{\textsc{Locater}}}
    \end{center}
    \caption{\textbf{Visual comparison results} {(\S\ref{sec:qua_vis}) on A2D-S \texttt{test}~\cite{gavrilyuk2018actor} (left) and R-YTVOS \texttt{val}~\cite{seo2020urvos} (right). Referents and corresponding descriptions are highlighted in the same colour.
    In right column, we show qualitative results of CSTM~\cite{hui2021collaborative} ({1\textit{st}}  row) and \textsc{Locater} ({2\textit{nd}} row).}}
    \label{fig:quali}
\end{figure*}

\noindent\textbf{Quantitative Performance.}
Following the official leaderboard, we report experimental results of \textsc{Locater} along with the SOTA LVS models~\cite{seo2020urvos,hui2021collaborative,li2022you} and two concurrent works~\cite{wu2022language,botach2022end} in Table~\ref{tab:youtubesota}. We observe that \textsc{Locater}  still performs well on this large-scale dataset. \textsc{Locater} consistently outperforms these methods on all metrics with the new SOTA scores, \ie, \bpo{56.5}/ \bpo{54.8}/\bpo{58.1} $\Jcal\&\Fcal$/$\Jcal$/$\Fcal$.

\subsection{Qualitative$_{\!}$ Analysis$_{\!}$ on$_{\!}$ A2D-S$_{\!}$ and$_{\!}$ R-YTVOS}\label{sec:qua_vis}
Fig.~\ref{fig:quali} depicts visual comparison results on A2D-S~\texttt{test} (left) and Refer-Youtube-VOS \texttt{val} (right).
\textsc{Locater} produces more precise segmentation results against ACGA~\cite{wang2019asymmetric} and CSTM~\cite{hui2021collaborative}.
It shows strong robustness in handling occlusions and complex textual descriptions,
especially when facing ambiguity caused by scene dynamics, where the referent is hard to locate relying on narrowly local perspective.
Particularly as shown in Fig.~\ref{fig:quali} (left), beyond the \textit{jump} action, ACGA (1\textit{st} row, 3\textit{rd} column) and CSTM (3\textit{rd} row, 4\textit{th} column) both fail to distinguish the two actors.

\subsection{Results on Our A2D-S$^+$ Dataset}\label{sec:contra_setting}

\noindent\textbf{Dataset.} Basically, {A2D-S}$^+_{\text{M}}$, {A2D-S}$^+_{\text{S}}$, and {A2D-S}$^+_{~\!\text{T}}$ yield increased challenges by collecting or synthesizing videos with multiple objects/actors. For {A2D-S}$^+_{\text{M}}$, it is a subset of A2D-S \texttt{test}, which contains \cnum{257} videos with multiple actors. For {A2D-S}$^+_{\text{S}}$ and {A2D-S}$^+_{~\!\text{T}}$, they are built upon the concept of \textit{contrasting} examples~\cite{sadhu2020video} (\cf\S\ref{sec:dataset_consturction}), \ie, videos that are similar to but not exactly the same as described by the language query. They contain \cnum{1290}/\cnum{1293} distinct videos with \cnum{1295}/\cnum{1295} linguistic queries respectively.

\noindent\textbf{Quantitative Performance.} Our A2D-S$^+$ dataset is only used for evaluation. All the benchmarked methods are the ones in Table~\ref{tab:a2dsota} trained on A2D-S~\cite{gavrilyuk2018actor} \texttt{train} set. As shown in Table~\ref{tab:subsets}, although \cite{wang2019asymmetric, hui2021collaborative} obtain compelling performance on A2D-S and J-HMDB-S (\cf~Table~\ref{tab:a2dsota}-\ref{tab:jhmdbsota}), they cannot handle our challenging cases well.
In contrast, our \textsc{Locater} yields better overall performance, especially on mIoU (\bpo{3.2} averaged improvement over all subsets), verifying its strong ability in fine-grained visual-linguistic understanding.

\noindent\textbf{Performance over Varied Number of Objects.}
Furthermore, for a thorough evaluation, we study the model performance with varied numbers of objects.
For either A2S-S$^+_\text{S}$ or A2S-S$^+_\text{T}$, we group the generated examples into three clusters according to the number of existing objects, and report the benchmarked results in Table~\ref{tab:num_objects}. The performance on the single actor subset of A2D-S \texttt{test} is also provided for reference.
As seen, as the number of semantically similar objects increases, the dataset difficulty also gradually elevated. Yet, the performance gap between \textsc{Locater} and other competitors becomes even larger, which well verifies the strong grounding ability of \textsc{Locater}.

\begin{table}[t]
    \centering
    \caption{\textbf{Quantitative results} {on $\text{A2D-S}^+$. \textsc{Locater}\sub{IMG} refers to \textsc{Locater} without using memory (\S\ref{sec:contra_setting}).}}
    \hspace{-1.5em}
    \tablestyle{0.5}{2}{1.1}{
        \begin{tabular}{|r||ccc|ccc|ccc|}
            \hline\thickhline
            \rowcolor{mygray}
                                                       & \multicolumn{3}{c|}{{A2D-S}$^+_{\text{M}}$} & \multicolumn{3}{c|}{{A2D-S}$^+_{\text{S}}$} & \multicolumn{3}{c|}{{A2D-S}$^+_{~\!\text{T}}$}                                                                                                 \\
            \rowcolor{mygray}
            \multicolumn{1}{|c||}{\multirow{-2}{*}{\textbf{Method}}}          & mAP   & oIoU   & mIoU  & mAP  & oIoU & mIoU & mAP  & oIoU & mIoU          \\
            \hline \hline
            \subt{19}{21}{\cite{wang2019asymmetric}}{~\sub{ICCV19}}   & 16.0  & 45.3    & 37.2 & 9.7   & 25.3 & 20.1 & 11.1 & 29.8 & 24.4          \\
            \subt{19}{21}{\cite{hui2021collaborative}}{~\sub{CVPR21}} & 26.7  & 50.6   & 43.3  & 16.3  & \textbf{34.1} & 28.7  & 19.2  & \textbf{38.3} & 27.5          \\
            \hline
            {\textsc{Locater}\sub{IMG}}          & 31.0 & 52.8  & 46.5 & 17.8 & 30.0 & 28.4 & 23.0 & 35.9 & 29.7          \\
            {\textbf{\textsc{Locater}}} & \textbf{33.6} & \textbf{54.5}  & \textbf{49.0} & \textbf{19.0} & 31.4 & \textbf{29.9} & \textbf{23.6} & 37.0 & \textbf{30.2} \\
            \hline
        \end{tabular}
    }
    \label{tab:subsets}
\end{table}

\begin{table}[t]
    \centering
    \caption{
        \textbf{Quantitative results} with different number of objects. Results are reported in terms of mIoU.
        \textsc{Locater}\sub{IMG} refers to \textsc{Locater} without using memory (\S\ref{sec:contra_setting}).
        }
    \hspace{-1.5em}
    \tablestyle{0.5}{3}{1.1}{
        \begin{tabular}{|r||c|cc|cc|cc|}
            \hline\thickhline
            \rowcolor{mygray}
            \#Objetcs/vid & {1} & 2.40 & 2.39 & 4.11 & 4.19 & 8.56 & 8.79 \\
            \rowcolor{mygray}
            Concatenation &  ~~~{-}~~~ & {Spat-} & {Temp-} & {Spat-} & {Temp-} & {Spat-} & {Temp-}  \\
            \hline \hline
            \subt{19}{21}{\cite{wang2019asymmetric}}{~\sub{ICCV19}}   & 69.8 & 29.9 & 34.2 & 19.5 & 25.0 & 11.1 & 13.9 \\
            \subt{19}{21}{\cite{hui2021collaborative}}{~\sub{CVPR21}} & 77.0 & 40.0 & 36.9 & 28.8 & 27.7 & 17.2 & 17.4 \\
            \hline
            {\textsc{Locater}\sub{IMG}} & 76.5 & 39.7 & 37.6 & 28.2 & 29.9 & 17.1 & 21.4 \\
            \textbf{\textsc{Locater}}   & 77.9 & 40.8 & 38.1 & 29.8 & 30.4 & 19.2 & 22.2 \\
            \hline
        \end{tabular}
    }
    \label{tab:num_objects}
\end{table}

\begin{table}[t]
	\centering
	\caption{\textbf{Benchmarking results} on \texttt{test-dev} set of RVOS track in YTB-VOS\sub{21} challenge$_{\!}$~\cite{vosc2021}~(\S\ref{sec:challenge}). \label{tab:challenge:val}}
	
	\resizebox{0.465\textwidth}{!}{
		\setlength\tabcolsep{6pt}
        \renewcommand\arraystretch{1.1}
		\begin{tabular}{|r||ccc|}
			\hline\thickhline
			\rowcolor{mygray}
			\textbf{Team} & $\mathcal{J}$\&$\mathcal{F}\uparrow$ & $\mathcal{J}\uparrow$ & $\mathcal{F}\uparrow$\\
			\hline \hline
			\textbf{LOCATER (Ours)} & \void{15}{10}{59.5} & \void{15}{10}{57.8} & {\void{15}{10}{61.1}} \\
			\textbf{leonnnop (Ours)} & \bbetter{15}{10}{\textbf{61.4}}{6.6}  & \bbetter{15}{10}{\textbf{60.0}}{6.3} & {\bbetter{15}{10}{\textbf{62.7}}{6.7}} \\
			\hline
			nowherespyfly~\cite{ding2021progressive} & \void{15}{10}{54.8} & \void{15}{10}{53.7} & {\void{15}{10}{56.0}} \\
			seonguk~\cite{seo2020urvos}              & \void{15}{10}{48.9} & \void{15}{10}{47.0} & {\void{15}{10}{50.8}} \\
			wangluting                               & \void{15}{10}{48.5} & \void{15}{10}{47.1} & {\void{15}{10}{49.9}} \\
			\hline
		\end{tabular}
	}
\end{table}

\begin{table}[t]
	\centering
	\caption{\textbf{Benchmarking results} on \texttt{test-challenge} set\protect\footnotemark of RVOS track in YTB-VOS\sub{21} challenge$_{\!}$~\cite{vosc2021}~(\S\ref{sec:challenge}). \label{tab:challenge}}
	
	\resizebox{0.465\textwidth}{!}{
		\setlength\tabcolsep{6pt}
        \renewcommand\arraystretch{1.1}
		\begin{tabular}{|r||ccc|}
			\hline\thickhline
			\rowcolor{mygray}
			\textbf{Team} & $\mathcal{J}$\&$\mathcal{F}\uparrow$ & $\mathcal{J}\uparrow$ & $\mathcal{F}\uparrow$\\
			\hline \hline
			\textbf{leonnnop (Ours)} & \bbetter{15}{10}{\textbf{60.7}}{11.3}  & \bbetter{15}{10}{\textbf{59.4}}{11.0} & {\bbetter{15}{10}{\textbf{62.0}}{11.7}} \\
			\hline
			nowherespyfly~\cite{ding2021progressive} & \void{15}{10}{49.4} & \void{15}{10}{48.4} & {\void{15}{10}{50.3}} \\
			feng915912132 & \void{15}{10}{48.2} & \void{15}{10}{47.4} & {\void{15}{10}{49.0}} \\
			Merci1        & \void{15}{10}{41.2} & \void{15}{10}{40.6} & {\void{15}{10}{41.8}} \\
			\hline
		\end{tabular}
	}
\end{table}
\footnotetext{\url{https://youtube-vos.org/challenge/2021/leaderboard/}}

\newcommand{\st}[4]{
    \grouptablestyle{1pt}{1}
    \begin{tabular}{d{#1}b{#2}}
    {#3} & {#4}
    \end{tabular}
}

\begin{table*}[t]
    \caption{\textbf{A set of ablation studies} (\S\ref{sec:ablation}) on A2D-S \texttt{test} and A2D-S$^{+}$. We report the average score on all three subsets of A2D-S$^+$.}
    \addtocounter{table}{-1} 

    \hspace{-0.7ex}
    \subfloat[{Key components}\label{tab:ablation:keycomponent}]{
        \resizebox{0.37\textwidth}{!}{
            \sgrouptablestyle{4pt}{1.07}
            \begin{tabular}[t]{|ccc||cc|}
                \hline\thickhline
                \rowcolor{mygray}
                $\Ecal_{vw}$     & $\Mcal$            & DSL              & \multicolumn{2}{c|}{\textbf{mIoU}}        \\
                \rowcolor{mygray}
                (\S\ref{sec:3.2.1}) & {(\S\ref{sec:3.2.2})} & (\S\ref{sec:3.2.3}) & A2D-S & A2D-S$^+$ \\
                \hline
                \hline
                                 &                    &                     & 47.3 & 27.0 \\
                \ding{51}        &                    &                     & 54.6 & 30.8  \\
                \ding{51}        & \ding{51}          &                     & 57.5 & 35.0 \\
                \ding{51}        &                    & \ding{51}           & 57.2 & 34.8 \\
                \ding{51}        & \ding{51}          & \ding{51}           & \textbf{59.7} & \textbf{36.4} \\
                \hline
            \end{tabular}
        }
    }\hspace{-1.1ex}
    \subfloat[{Cross-modality encoder}\label{tab:ablation:cme}]{%
        \resizebox{0.267\textwidth}{!}{
            \sgrouptablestyle{6pt}{1.02}
            \begin{tabular}[t]{|c||cc|}
                \hline\thickhline
                \rowcolor{mygray}
                $\Ecal_{vw}$     & \multicolumn{2}{c|}{\textbf{mIoU}}                 \\
                \rowcolor{mygray}
                (\S\ref{sec:3.2.1}) & A2D-S & A2D-S$^+$          \\
                \hline
                \hline
                K=1                             & 58.4          & 35.2 \\
                K=2                             & 59.0          & 35.8 \\
                K=3                             & \textbf{59.7} & \textbf{36.4} \\
                K=4                             & 59.5          & 36.4 \\
                K=5                             & 59.7          & 36.4 \\
                \hline
            \end{tabular}
        }
    }\hspace{-1.1ex}
    \subfloat[{Local-global memory (mIoU on A2D-S$^+$)}\label{tab:ablation:lgmemory}]{%
        \resizebox{0.34\textwidth}{!}{
            \sgrouptablestyle{3pt}{1.05}
            \begin{tabular}[t]{|c||ccccc|}
                \hline\thickhline
        \rowcolor{mygray}
                \cellcolor{mygray}\backslashbox{$N_g$}{$N_l$} &\cellcolor{mygray} 0 &\cellcolor{mygray} $N_v$ &\cellcolor{mygray} 1.5$N_v$ &\cellcolor{mygray} 2$N_v$ &\cellcolor{mygray} 3$N_v$ \\
                \hline
                \hline
                0             & 34.8 & 35.2 & 35.4 & 35.4 & 35.5 \\
                $N_v$         & 35.3 & 35.5 & 35.8 & 36.0 & 36.1   \\
                1.5$N_v$      & 35.3 & 35.6 & 36.1 & \textbf{36.4} & 36.4 \\
                2$N_v$        & 35.4 & 35.8 & 36.2 & 36.4 & 36.4   \\
                3$N_v$        & 35.4 & 35.9 & 36.2 & 36.4 & 36.4   \\
                \hline
            \end{tabular}
        }
    }\vfill\vspace{-.03in}
    \subfloat[{Visual encoder}\label{tab:ablation:ve}]{%
        \resizebox{0.31\textwidth}{!}{
            \sgrouptablestyle{3pt}{1}
            \begin{tabular}[t]{|x{58}||x{30}x{30}|}
                \hline\thickhline
                \rowcolor{mygray}
                $\Ecal_{v}$                &  \multicolumn{2}{c|}{\textbf{mIoU}}                 \\
                \rowcolor{mygray}
                {(\S\ref{sec:3.2.1})}      & A2D-S                  & A2D-S$^+$          \\
                \hline
                I3D~\cite{carreira2017quo} & 58.3          & 35.2 \\
                Transformer                & \textbf{59.7} & \textbf{36.4} \\
                \hline
            \end{tabular}
        }
    }\hspace{-2.1ex}
    \subfloat[{Contextualized query embedding}\label{tab:ablation:queryemb}]{%
        \resizebox{0.37\textwidth}{!}{
            \sgrouptablestyle{3pt}{1.025}
            \begin{tabular}[t]{|x{90}||x{30}x{30}|}
                \hline\thickhline
                \rowcolor{mygray}
                Query embedding $\qbm$ & \multicolumn{2}{c|}{\textbf{mIoU}}                 \\
                \rowcolor{mygray}
                (\S\ref{sec:3.2.3})    & A2D-S                          & A2D-S$^+$          \\
                \hline
                Text-invented          & 58.7          & 35.7 \\
                Visual-guided          & \textbf{59.7} & \textbf{36.4} \\
                \hline
            \end{tabular}
        }
    }\hspace{-2.1ex}
    \subfloat[{Decoding mode (mIoU)}\label{tab:ablation:qm}]{%
        \resizebox{0.32\textwidth}{!}{
            \sgrouptablestyle{3pt}{1.05}
            \begin{tabular}[t]{|x{70}||x{30}x{30}|}
                \hline\thickhline
                \rowcolor{mygray}
                Mode (Eq.~\ref{eq:readout}) & A2D-S & A2D-S$^+$          \\
                \hline
                Encoder only            & 56.8          & 34.1 \\
                Cosine distance         & 59.2          & 36.3 \\
                Product distance        & \textbf{59.7} & \textbf{36.4} \\
                \hline
            \end{tabular}
        }
    }\vfill\vspace{-.03in}
    \subfloat[{Efficiency comparison}\label{tab:ablation:efficiency}]{%
        \resizebox{0.495\textwidth}{!}{
            \sgrouptablestyle{5pt}{1.}
            \begin{tabular}{|l||ccccx{55}|}
                \hline\thickhline
                \rowcolor{mygray}
                \multicolumn{1}{l||}{$\#$ Frame} & 15 & 30 & 50 & 80 & {\st{15}{36}{100}{}} \\
                \hline\hline
                Transformer-style          & 2.01 & 2.95 & 4.57 & 8.48 & {\st{15}{36}{11.54}{(\textbf{ms/f})}} \\
                \textsc{Locater} (Memory)  & \textbf{1.71} & \textbf{2.00} & \textbf{2.22} & \textbf{2.26} & {\st{15}{36}{\textbf{2.26}}{(\textbf{ms/f})}} \\
                \hline
            \end{tabular}
        }
    }\hspace{-1.6ex}
    \subfloat[{Frame sampling interval (mIoU)}\label{tab:ablation:sample_rate}]{%
        \resizebox{0.495\textwidth}{!}{
            \sgrouptablestyle{4pt}{1}
            \begin{tabular}{|l||cccccccc|}
                \hline\thickhline
                \rowcolor{mygray}
                Interval ($\#$Frame) &3  &5 & 7 & 10 & 12 & 15 & 20 & 30 \\\hline\hline
                ~~~~~~~A2D-S     & 59.7 & 59.7 & 59.7 & 59.7 & 59.6 & 59.4 & 59.3 & 59.0 \\
                ~~~~~~~A2D-S$^+$ & 36.4 & 36.4 & 36.4 & 36.4 & 36.4 & 36.1 & 35.9 & 35.5 \\
                \hline
            \end{tabular}
        }
    }
\end{table*}

\subsection{Results on RVOS Track in YTB-VOS\sub{21} Challenge}\label{sec:challenge}

\noindent\textbf{Experimental Setup.} We first detail the experimental setup for the challenge dataset in YTB-VOS\sub{21}.
\begin{itemize}[leftmargin=*]
	\setlength{\itemsep}{0pt}
	\setlength{\parsep}{1pt}
	\setlength{\parskip}{1pt}
	\setlength{\leftmargin}{-10pt}
    \item \textit{Dataset}: R-YTVOS~\cite{seo2020urvos} (\cf\S\ref{sec:urvos}) is the standard benchmark. Challenge solutions are first developed on the \texttt{test-dev} set, which contains the same video sequences as R-YTVOS \texttt{val}, and finally evaluated on the private \texttt{test-challenge} set, which contains \cnum{305} videos.
    \item \textit{Evaluation Metric}: Following the official evaluation metrics, we use $\Jcal\&\Fcal$, $\Jcal$ and $\Fcal$ to evaluate our model.
    \item \textit{Implementation Details}: We resize video frames to $384\!\x\!384$ and separate them with a patch size $O\!=\!8$, which results in $N_v\!=\!2,304$. The linguistic encoder $\Ecal_{w}$ is implemented as BERT\sub{BASE}~\cite{devlin2018bert}. Our model is trained with an initial learning rate \cnum{2e-4} which decays polynomially~\cite{liang2022gmmseg}, batch size \cnum{48}, weight decay \cnum{1e-4} and max epoch \cnum{50}. During testing, we adopt multi-scale inference with horizonal flip and scales of $[0.5, 0.75, 1.0, 1.25, 1.5, 1.75]$.
    Other settings are kept unchanged (\cf\S\ref{sec:detail}).
\end{itemize}

\noindent\textbf{Modifications.}
In addition to the common challenges in LVS benchmarks, we observe a unique issue in YTB-VOS\sub{21}:
A part of objects in the \texttt{test-challenge} set belongs to novel categories that are unseen in the \texttt{train} set.
Thus, along with the \textsc{Locater} trained on YTB-VOS\sub{21} challenge dataset, we further ensemble other RES models, \ie, MCN~\cite{luo2020multi}, which are trained on open-set data sources, including image-level datasets like RefCOCO~\cite{yu2016modeling}, RefCOCOg~\cite{yu2016modeling} and RefCOCO+~\cite{mao2016generation}.
To effectively combine these predictions, we propose a standalone grounding module.
Specifically, the grounding module takes several sequence-level mask predictions as inputs and predicts the similarity scores based on the referring expression. It is implemented as four stacked Transformer blocks~\cite{vaswani2017attention} followed by one MLP layer for score prediction.
This module complements the precise segmentation masks from \textsc{Locater} with novel yet coarse object masks from other models.

\noindent\textbf{Quantitative Results.}
In Table~\ref{tab:challenge:val} and~\ref{tab:challenge}, we respectively report the final results of our final solution and other top-leading teams on the \texttt{test-dev} and \texttt{test-challenge} sets of the RVOS track in YTB-VOS\sub{21}. For completeness, in Table~\ref{tab:challenge:val}, we also report the performance of \textsc{Locater} trained under the same setup to our full solution.
Other competitors~\cite{ding2021progressive} mainly adopt an image-level referring object grounding strategy and simply generate video-level predictions with a fixed tracking module.
They not only neglect the indispensable long-term cues within linguistic expressions but also overlook the intrinsic low-level challenges posed within video sequences; In contrast, our model well-addresses these issues.
As seen, our final solution significantly surpasses the second-place solution with a large gap of \bpo{11.3}/\bpo{11.0}/\bpo{11.7} in terms of $\Jcal\&\Fcal$/$\Jcal$/$\Fcal$ on \texttt{test-challenge} set. Notably, \textsc{Locater} has already surpassed all other competitors significantly.

\subsection{Diagnostic Experiments}\label{sec:ablation}
In this section, we conduct a series of ablative studies on both A2D-S \texttt{test}~\cite{gavrilyuk2018actor} and our newly proposed A2D-S$^+$ to fully examine the efficacy of our algorithm design. For A2D-S$^+$, the score is reported as the average over all the three subsets, \ie, A2D-S$^+_{\text{M}}$, A2D-S$^+_{\text{S}}$ and A2D-S$^+_{\text{T}}$.

\noindent\textbf{Key Component Analysis.}
To study the effect of essential components of \textsc{Locater}, we first establish a baseline, that only remains single-modal encoders and directly concatenates visual and linguistic features~\cite{he2016deep,li2017tracking,gavrilyuk2018actor} for mask prediction (\ie, the first row in Table~\ref{tab:ablation:keycomponent}). Then we gradually add different modules, \ie, \textit{cross-modality encoder} ($\Ecal_{vw}$), \textit{local-global memory} ($\Mcal$), and \textit{deeply-supervised learning} (DSL) strategy, into the baseline. As reported in Table~\ref{tab:ablation:keycomponent}, all these components indeed boost segmentation and combining them together yields the best performance.

\noindent\textbf{Visual Encoder $\Ecal_{v}$.} Next, to verify the advantage of our fully attentional model design, we replace Transformer based visual encoder $\Ecal_{v}$ with traditional I3D~\cite{carreira2017quo}, the conventional backbone in previous works~\cite{wang2019asymmetric,wang2020context,hui2021collaborative,yang2021hierarchical}, and observe performance degradation in Table~\ref{tab:ablation:ve}. We further notice that, even with I3D as visual backbone, \textsc{Locater} still outperforms existing algorithms, if we compare the results with Table~\ref{tab:a2dsota}, \ie, \po{58.3} \textit{vs} \po{56.1} (+\po{2.2}) in terms of mIoU.

\noindent\textbf{Cross-Modality Encoder $\Ecal_{vw}$.} Then we study the efficacy of our cross-modality encoder $\Ecal_{vw\!}$ design, which has $K_{\!}\!=_{\!}\!3$ attention-based modules. As shown in Table~\ref{tab:ablation:cme}, the performance increases when stacking more modules ($K{\!}:_{\!} 1\!\rightarrow\!3$), and then the gain becomes marginal ($K{\!}:_{\!} 3\!\rightarrow\!5$).

\noindent\textbf{Decoding Mode} (Eq.~\ref{eq:readout}).
We further study the influence of different decoding modes on performance.
We consider two variants:
\textbf{i)} \textit{encoder only} (\ie, replace the decoder~with~a MLP based segment head); and
\textbf{ii)} \textit{cosine distance} (\ie, compute the scalar product between $\ell_2$-normalized query and patch embeddings:).
As shown in Table~\ref{tab:ablation:qm}, directly removing the decoder triggers a clear performance drop (\po{59.2}$\rightarrow$\po{56.8} on A2D-S \texttt{test} in terms of mIoU). Compared with \textit{cosine distance}, we find product distance is more favored: \po{59.7} \textit{vs} \po{59.2}. This observation is reconfirmed on A2D-S$^+$.

\begin{figure*}[t]
	\begin{center}
		\includegraphics[width=1\linewidth]{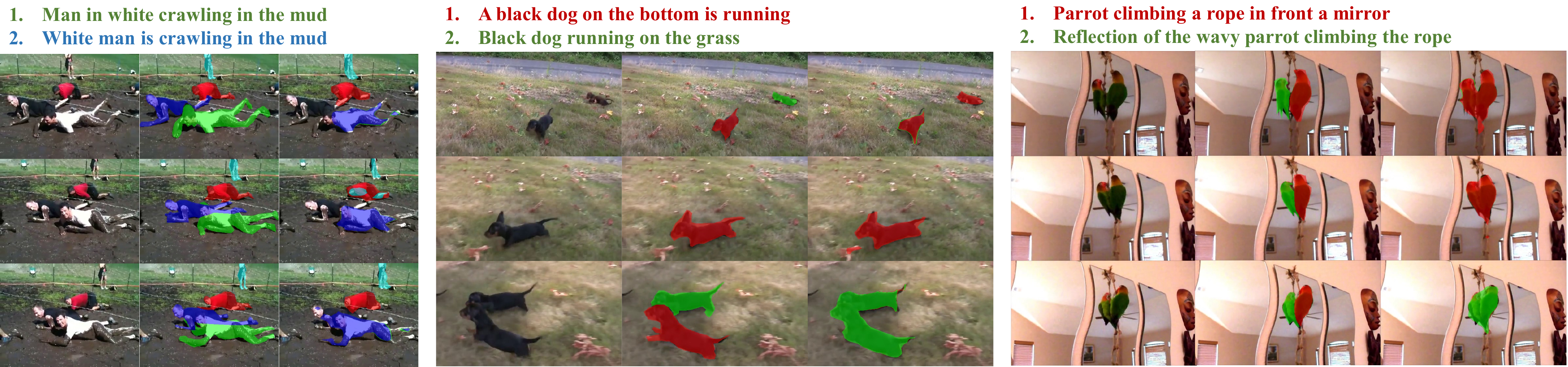}
        \put(-518, -6){\scalebox{.70}{{Input Sequence}}}
        \put(-469, -6){\scalebox{.70}{{Ground Truth}}}
        \put(-418, -6){\scalebox{.70}{\textsc{Locater}}}
        \put(-365, -6){\scalebox{.70}{{Input Sequence}}}
        \put(-301, -6){\scalebox{.70}{{Ground Truth}}}
        \put(-233, -6){\scalebox{.70}{\textsc{Locater}}}
        \put(-175, -6){\scalebox{.70}{{Input Sequence}}}
        \put(-112, -6){\scalebox{.70}{{Ground Truth}}}
        \put(-45,  -6){\scalebox{.70}{\textsc{Locater}}}
	\end{center}
	\caption{Typical failure cases on A2D-S \texttt{test}. See \S\ref{sec:discussion} for more details.}
    \label{fig:fail}
\end{figure*}
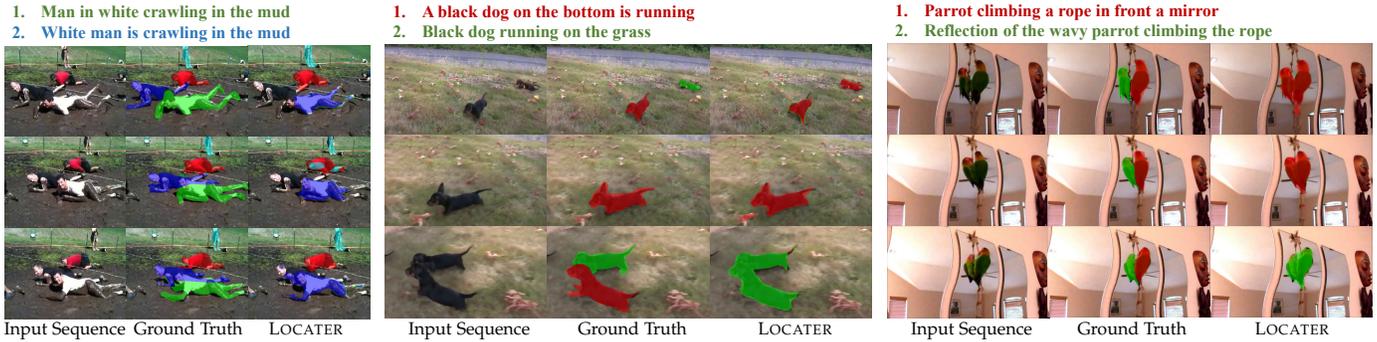

\noindent\textbf{Local-Global Memory $\Mcal$.} Table$_{\!}$~\ref{tab:ablation:lgmemory} summarizes the impact of the capacity of the local-global memory $\Mcal$ in segmentation. As seen, without using either local ($N_l\!=\!0$) or global memory ($N_g\!=\!0$), or even both ($N_l\!=\!0$ and $N_g\!=\!0$), \textsc{Locater} suffers from significant performance drop. With the increase of the local or global memory capacity, the performance is improved but the gain becomes less stark, confirming the value of the learnable memory operations.

\noindent\textbf{Contextualized Query Embedding.} In \S\ref{sec:3.2.3}, we generate a compact query embedding $\qbm$ based on visual context (Eq.$_{\!}$~\ref{eq:query}). To investigate such a design, we consider an alternative strategy, \ie, applying self-attention to generate $\qbm$. As in$_{\!}$ Table$_{\!}$~\ref{tab:ablation:queryemb},$_{\!}$ visual-guided$_{\!}$ query$_{\!}$ embedding$_{\!}$ is$_{\!}$ more$_{\!}$ favored.

\noindent\textbf{Efficiency of Memory Design.}
We further demonstrate the efficiency of the finite memory design in Table~\ref{tab:ablation:efficiency}.
To clearly reveal the gap, we directly feed image features to both one Transformer block~\cite{vaswani2017attention} with 12 heads and our memory module (\cf~\S\ref{sec:3.2.2}). Every frame is patch-wise separated into \cnum{256} tokens with \cnum{768} feature channels for each.
Compared with using conventional Transformer-style network design, \ie, flattening all tokens as input, the advantage of our method in efficiency is more and more significant with the increase of video length.

\noindent\textbf{Frame Sampling Interval.} Moreover, we evaluate the impact of frame sampling interval used during global memory construction (\S\ref{sec:3.2.2}).
As shown in Table~\ref{tab:ablation:sample_rate}, sampling more frames with smaller intervals ($\leq$10) cannot provide performance gain, showing the high redundancy among video frames and explaining the high efficiency of our approach. We, therefore, set the default sampling interval as \cnum{10} to construct the global memory $\Mcal^{g}$.

\subsection{Failure Case Analysis}\label{sec:discussion}
Despite the stronger cross-modal video understanding ability of \textsc{Locater} against previous methods, it still suffers from difficulties in some challenging scenarios.
We present three typical failure modes on A2D-S \texttt{test} in Fig.~\ref{fig:fail}.
The \textit{first} type of mistakes is caused by the ambiguity of natural language. For example, in $1\textit{st}$ column, \textsc{Locater} with an LSTM-based language model, which is trained from scratch, is hard to distinguish the two reference, \ie, \textit{white man} and \textit{man in white}, with limited language training data.
This issue can be alleviated by introducing a large-scale pretrained language model, \eg, BERT~\cite{devlin2018bert}, or harvesting finer language modeling.
The \textit{second} type of challenges comes from weak or ambiguous descriptions. For example, in $2\textit{nd}$ column, the referent cannot be well differentiated from other objects, as both of the two black dogs are matched well with the query ``\textit{Black dog running on the grass}''.
The \textit{third} type of challenges is brought by highly similar referents. For example, in $3\textit{rd}$ column, our model faces difficulties in distinguishing the parrot and its reflection, as their appearance and motion information are almost exactly the same.

\section{Conclusion}

This work presents a memory augmented, fully attentional model, \textsc{Locater}, for LVS. It effectively aligns cross-modal representations, and efficiently models long-term temporal context as well as short-term segmentation history through an external memory. With visual context guided expression attention, \textsc{Locater} produces frame-specific query vectors for mask generation.
We further observe and mitigate the critical absence issue of grounding-required objects in the current most popular A2D-S benchmark {$_{\!}$}with {$_{\!}$}a {$_{\!}$}newly {$_{\!}$}created A2D-S$^+$ dataset. {$_{\!}$}It, {$_{\!}$}as {$_{\!}$}a {$_{\!}$}sufficient {$_{\!}$}complement, significantly increases the number of semantically similar objects in testing examples.
Experiments demonstrate that \textsc{Locater} dramatically advances state-of-the-arts with high efficiency on both standard benchmarks and our proposed challenging A2D-S$^+$.
Furthermore, our \textsc{Locater} based solution achieved the 1\textit{st} place in RVOS Track of YTB-VOS\sub{21} Challenge, surpassing other competitors by large margins.

\ifCLASSOPTIONcaptionsoff
  \newpage
\fi

{\small
\bibliographystyle{IEEEtran}
\bibliography{egbib}
}

\end{document}